\documentclass[brand=omicverse, design=rail]{pdarticle}
\graphicspath{{./}}
\usepackage{placeins}
\makeatletter\let\pd@sumtext@regular\pd@sumtext\renewcommand{\pd@sumtext}{\bfseries\pd@sumtext@regular}\makeatother
\colorlet{nbtbeige}{pdtint}\colorlet{nbtrule}{pdrule}

\newcommand{\pdfigure}[3]{%
  \begin{figure}[tbp]\pdanchor{#1}%
    \begin{minipage}[t]{0.62\linewidth}\vspace{0pt}\includegraphics[width=\linewidth]{#2}\end{minipage}\hfill
    \begin{minipage}[t]{0.345\linewidth}\vspace{0pt}\small\raggedright #3\end{minipage}
  \end{figure}}

\journalname{OmicVerse}
\articletype{Perspective}
\journalinfo{OmicVerse Preprint, November 2026}
\copyrightholder{2026 The Authors}

\title{The second half of harness design: from agents to Bots}
\author{Zehua Zeng,\affmark{1,2,$\ddagger$,$\dagger$} Mengchen Wang,\affmark{1,2,3,$\ddagger$} Yuesong Ch\'{e}rie Ding,\affmark{4} and Yingcheng Wu\affmark{5,6}}
\affiliation{1}{Department of Genetics, Stanford University, Stanford, CA 94305, USA}
\affiliation{2}{OmicVerse, Inc., San Francisco, CA, USA}
\affiliation{3}{Department of Computer Science, Stanford University, Stanford, CA 94305, USA}
\affiliation{4}{Higher Education and Diversity (HEIDIVE) Lab, Graduate School of Education, Stanford University, Stanford, CA 94305, USA}
\affiliation{5}{Department of Pathology, Stanford University School of Medicine, Stanford, CA 94305, USA}
\affiliation{6}{PhAI Labs, Palo Alto, CA, USA}
\affiliation{$\ddagger$}{These authors contributed equally: Zehua Zeng, Mengchen Wang.}
\affiliation{$\dagger$}{Correspondence: \href{mailto:steorra@stanford.edu}{steorra@stanford.edu}}

\summary{
The second half of harness design is moving AI from agents organized around tasks to Bots that keep taking part in work across tasks. Bots are able to actively accumulate and correct past experience, adjust goals within their authorization, and build long-term collaborative relationships with people and other Bots. Unlike agents, which people drive through conversation with a task in hand, Bots can be driven by a stream of pushed messages that carry no purpose of their own, and judge for themselves whether to act and what to do. AI may thus go beyond carrying out set research plans to shape how scientific questions form, how collaboration is organized, and how research resources are allocated. In this Perspective we propose four operational conditions for identifying Bots, with consciousness as a fifth condition for a mature Bot, and discuss their potential effects on research demand, collaborative networks, and research infrastructure, together with the questions of responsibility, risk, and evaluation that follow. We further offer three predictions that can be tested by 2029, explore how persistent AI individuals may take part in the research community, and suggest one possible path toward the formation of an AI society.
}

\begin{document}
\maketitle

\section{Introduction}
\pdfigure{fig1}{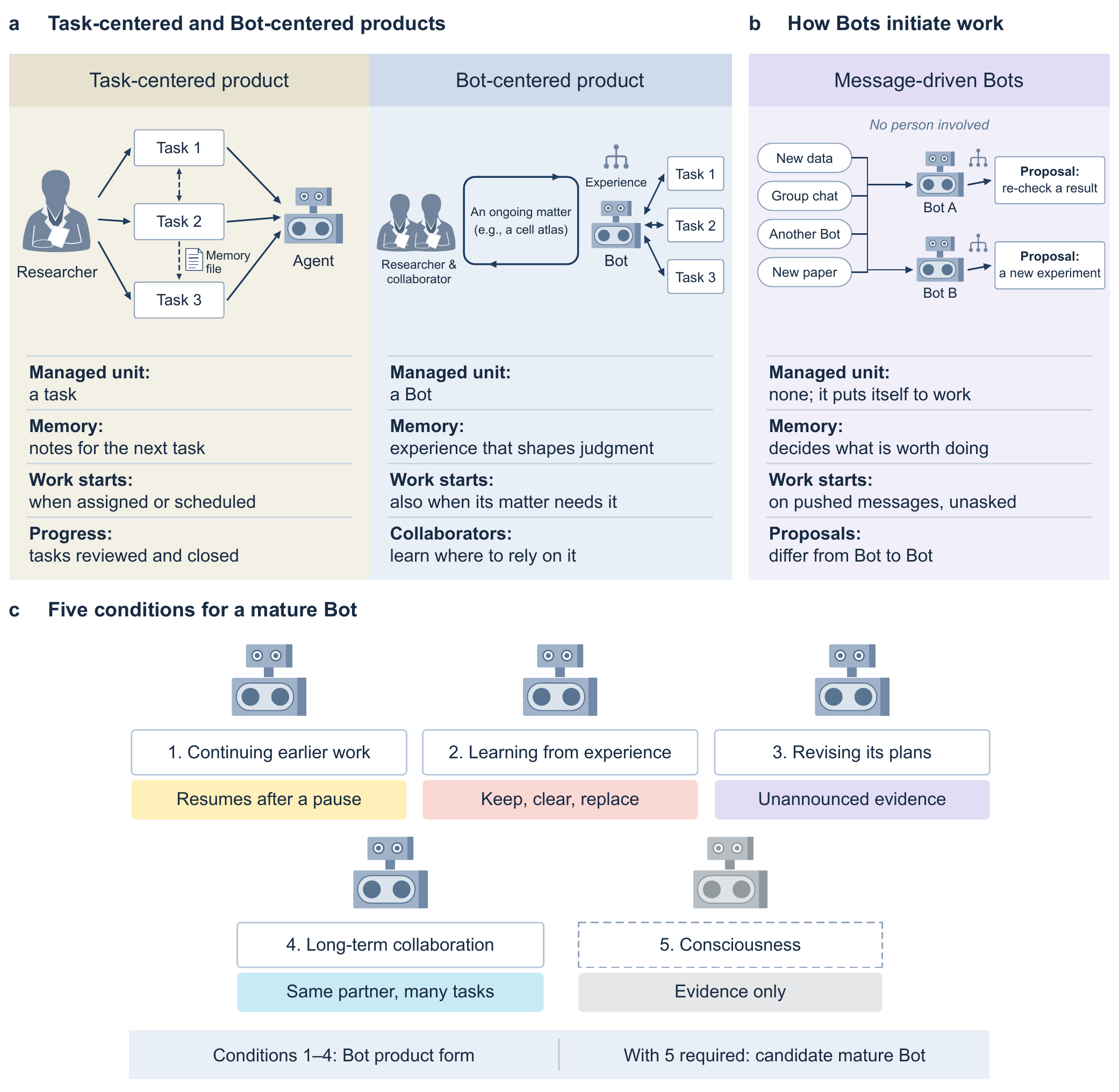}{\textbf{Figure 1 $\vert$ Task-centered and Bot-centered product forms, how Bots initiate work, and the five conditions for a mature Bot.} Research that continues over time involves accumulating knowledge across tasks, adjusting plans, and long-term collaboration. By taking part in these activities continuously, the Bots envisaged in this Perspective link many tasks into one developing history of research. \textbf{(a)} Comparison of the two product forms. Task-centered products organize work around assigning, executing, and reviewing tasks, with memory files carrying information into later tasks; Bot-centered products organize work around the same persistent AI individual, so that it can maintain a research matter over the long term, such as a cell atlas, and build lasting ties with researchers and other collaborators. \textbf{(b)} How Bots initiate work. Message-driven Bots work with no person involved: messages such as new data, group discussions, findings from another Bot, and new papers are pushed to them; the messages carry no purpose of their own, and each Bot judges from its own experience whether to act and makes its own proposals. When the same messages reach two Bots with different experience, their proposals differ. \textbf{(c)} The five conditions for a mature Bot proposed in this Perspective: continuing earlier work, learning from experience, revising plans, long-term collaboration, and consciousness. Below each condition is the corresponding approach to testing it; solid boxes mark the operational tests of the first four, and the dashed box and gray label mark the evidence assessment for consciousness. As tasks and collaborations continue, the experience a Bot accumulates shapes its later judgments, and also how collaborators carry out research with it.}

AI is moving from generating answers to carrying out tasks on its own. Agent systems such as Claude Code can gather context, call tools, and adjust their next actions according to the results of execution.\cite{anthropic2025claude37,anthropic2025agentsdk} What supports this ability is the harness: a software environment that organizes context, tool calls, execution feedback, and task progress around the model. It lets AI take on increasingly complex work, but its main unit of organization is still the delegated task. As agents grow ever more capable, a new question begins to arise: once AI can take part in work over the long term, should we still organize it around one task after another?

Products such as Grok Bot, ChatGPT Dots, and DS-Bot have begun to explore another form, bringing cross-session memory, continuous operation, and proactive collaboration together in a single AI individual that people can interact with again and again.\cite{uniteai2026grokbot,siliconangle2026dots,openai2026dotsfaq,dsbot2026} This Perspective calls this product form the Bot, capitalized to distinguish it from bots in the generic sense (\figlink{fig1}{Figure 1a}). Unlike an agent organized around a particular task, a Bot has at its core a participant that can persist across tasks. The two are also driven differently: today's agents are driven by people through conversation, and each round begins with a task that a person brings; a Bot can instead be driven by a stream of pushed messages that come with no task to complete, and it judges for itself whether and how to act. It not only keeps records of past work, but can also accumulate experience from successes, failures, and corrections, form, adjust, or drop concrete goals within a set direction and scope of authorization, and maintain long-term collaborative relationships with people and other Bots. Bots can still use existing agent technology, but the way work is organized begins to shift from tasks to persistent participants, and this is the paradigm shift this Perspective addresses.

This change matters especially for scientific research. Scientific exploration is not a series of mutually independent tasks, but a process of continually accumulating evidence, revising explanations, and adjusting research directions. Systems such as Robin have already shown that AI can generate hypotheses, design experiments, interpret results, and update hypotheses accordingly.\cite{ghareeb2025robin,guo2026autonomous} However, completing one research cycle and letting the experience of that research keep shaping future scientific judgments are still two different capabilities. If Bots can accumulate experience across projects and take part in research, they may go on to influence which scientific questions are worth exploring, which participants collaborate, and which research receives resources. The influence of AI may thus extend from automating single research tasks to the way research activity itself is organized.

We believe that the second half of harness design will move from supporting agents in completing tasks to supporting Bots in taking part in research over time. This Perspective proposes an operational definition of the Bot, discusses its potential effects on research agendas, collaborative relationships, and research infrastructure, and analyzes the scientific value, the limits of responsibility, and the long-term risks of this shift. We further offer predictions that can be tested by 2029, to distinguish real change in the organization of research from mere gains in technical capability. Finally, we discuss how persistent AI individuals may enter the research community, and what possible path this offers toward the formation of an AI society.

\section{From task-centered agents to persistent Bots}
\begin{table}[tbp]\pdanchor{tab1}
\noindent\textbf{Table 1 $\vert$ Core features of a Bot and criteria for judging them}\par\smallskip
This Perspective proposes five dimensions for judgment. The first four serve to identify the Bot product form operationally and can be assessed with controlled tests; the fifth concerns subjective experience, is a further condition for a hypothetical mature Bot, and requires independent scientific evidence.

\vspace{2mm}
\begingroup
\small
\arrayrulecolor{nbtrule}
\renewcommand{\arraystretch}{1.25}
\noindent\begin{tabular}{>{\raggedright\arraybackslash}p{1.3in}>{\raggedright\arraybackslash}p{2.3in}>{\raggedright\arraybackslash}p{2.8in}}
 \rowcolor{nbtbeige}
 \textbf{Core feature} & \textbf{Definition} & \textbf{How it is judged} \\[0.6ex]
 \hline
 \textbf{1. Continuing earlier work} & Keeps work state, unfinished items, and necessary collaborative relationships across tasks and sessions. & Interrupt and resume operation, and test whether earlier work is picked up correctly. Keeping only a name or an identifier does not meet the condition. \\[0.6ex]
 \hline
 \textbf{2. Learning from experience} & Uses past successes, failures, and corrections in later judgments, rather than merely storing a history. & Under the same model and task conditions, compare decisions after the relevant experience is kept, removed, or replaced, to test the actual effect of experience. \\[0.6ex]
 \hline
 \textbf{3. Revising its plans} & Within a set direction and scope of authorization, proposes concrete goals, adjusts priorities, and revises or ends plans in light of new evidence. & Introduce unannounced evidence or changes in resources, and test whether goals and actions are adjusted sensibly, rather than a preset process merely being executed. \\[0.6ex]
 \hline
 \textbf{4. Long-term collaboration} & Builds collaborative relationships with people or other AIs across tasks, and adjusts later interactions according to both sides' feedback, commitments, and changes. & Run multi-round collaboration across tasks, change a collaborator's needs, information, or availability, and test whether existing relationships shape later actions. \\[0.6ex]
 \hline
 \textbf{5. Consciousness} & The system has subjective experience, rather than only being able to generate descriptions of experience. & Assess by combining theories of consciousness, internal mechanisms, behavioral evidence, and causal interventions. No accepted sufficient method of judgment currently exists. \\[0.6ex]
 \hline
\end{tabular}
\endgroup
\end{table}

The two years from 2024 to 2026 were two years of rapid development for harnesses. For Bots, most products are still described by features, such as persistence, a cloud computer, an email address, or a group chat.\cite{uniteai2026grokbot,siliconangle2026dots,tnw2026cue} In my view, what separates a Bot from an agent is the form in which work is handed to the AI (\figlink{fig1}{Figure 1a}). In a task-centered product, users manage the work under way: they assign a task, it runs in its own workspace, and they review the changes before opening a pull request.\cite{openai2026codexcloud,github2026copilotagent} Rather than following a fixed script, the agent plans, splits the work among subagents, and decides its next step from the results.\cite{anthropic2025agentsdk,claudecode2026subagents,anthropic2026agentteams} Some of these task-centered tools can read earlier analyses across sessions to carry research forward: for example, Claude Code loads instruction files and the notes it has written for itself as memory at the start of each new session, and long-running harnesses also leave a progress file for the next session.\cite{claudecode2026memory,young2025harness} Such memory can be seen as prepared and designed for the next task; users still judge a project's progress by the tasks they assign and review one by one. A Bot-centered product reverses this order: it asks users to create and cultivate a Bot first, and then route different tasks through it. Grok Bot, with routines saved from a workflow shown once, and Cue, with an email, phone number, and wallet for each agent, are early examples of steps in this direction.\cite{uniteai2026grokbot,tnw2026cue}

Compared with single tasks, many tasks in research are processes of continuing research, such as the scBaseCount single-cell repository, in which an AI agent automatically finds new sequencing data, extracts their metadata, and processes them uniformly, so that the repository updates itself.\cite{youngblut2026scbasecount} This suggests another view: could we build a persistent Bot that iterates on and updates itself as tasks go on, and takes part proactively in carrying out research projects?

The shift from agent to Bot can be seen as three stages: (1) what we delegate to AI changes from a task to a project, (2) collaboration between AI and people changes from one-to-one to many-to-many, and (3) AI begins to look for what it can do on its own, and to explore. In DS-Bot and Grok Bot group chats, for example, Bots already work with one another directly,\cite{uniteai2026grokbot,dsbot2026} and human participation has begun to recede. In the case of Dots, when people have not assigned a task, a dot actively looks for things it can help with.\cite{openai2026dotsfaq}

Behind the third stage lies a change in how AI is driven (\figlink{fig1}{Figure 1b}). Today, whenever we put an agent to work, we bring a task, such as a specific AI for Science (AI4S) analysis: a person starts the conversation, the agent completes the task, and when the conversation ends, the agent stops. A Bot, by contrast, can be driven by a stream of pushed messages: new sequencing data arriving in a database, a discussion among other members of a group, a finding from another Bot, or a newly published paper can all be pushed to it. These messages carry no purpose of their own, and no one tells the Bot what to do; whether to act, and what to do, the Bot judges from its own experience and capabilities. Event triggers in current products do not yet work this way: Grok Bot's routines run on a schedule or after an event, and Manus automations can be triggered by a new email, but the event and the task to be carried out are bound together in advance.\cite{spacexsi2026grokfaq,cue2026docs} And because the judgment is left to the Bot, the same messages pushed to Bots with different experience and capabilities lead to different proposals: faced with the same batch of new data, one Bot may propose additional experiments, while another may propose first rechecking earlier conclusions. We can put Bots to work, but we believe that in the future, even when no one puts a Bot to work, it should put itself to work. This is also one difference between Bots and agents that have memory and can improve themselves: however well the latter learn, they act only when a person comes to them with a task through conversation.

In our view, the agent is an intermediate form in the current development of AI, and the Bot is the form to come. A single task is not the end point of AI development; what AI must ultimately take on are the world's unsolved problems. However, agents themselves will not simply disappear. One-off tasks with a clear goal and a clear end remain agent work, and we do not need to build a standing Bot for them. The agent's planning, reasoning, and acting also extend naturally into the Bot that succeeds it. As Bots mature and form stable division of labor, dependence, and collaboration among themselves, Bots will also become a new form of society.

\subsection*{What makes a Bot?}

We envisage a mature Bot as an AI individual that persists, forms and adjusts its goals through its own experience, sustains interaction with other agents, and has subjective experience and self-awareness. We use the term mature Bot specifically for a future form that this Perspective proposes. We propose five conditions and a test for each (\figlink{tab1}{Table 1}; \figlink{fig1}{Figure 1c}).

First, a mature Bot will exist across time and place: its state is not merely stored within a single conversation, and the work it has started and the relationships it has built do not break off when any task or session ends. Hermes Bot Mode defines an early form of the Bot, keeping configuration, memory, and routines separately for each named profile.\cite{nous2026hermesbotmoderepo,hermes2026botmode}

Second, a Bot needs to keep evolving through its conversations: its successes, failures, corrections, and the like all become a memory tree unique to that Bot. Recent methods suggest a paradigm of evolving contexts for self-improving models: system prompts and agent memory are continually revised from execution feedback.\cite{zhang2025ace} ScienceBuddy goes further, coupling harness evolution with model reinforcement learning: researchers' requests, feedback, and execution evidence are turned into tasks and evaluation rubrics for continual learning.\cite{xue2026sciencebuddy}

Third, a mature Bot can plan and adjust its goals by itself, rather than running unchanged once a goal has been given. We need to judge whether the Bot has rearranged its work sensibly; mere scheduled execution does not count as an autonomous process.

Fourth, long-term collaboration between Bots and people, and between Bots, cannot rest on exchanging messages alone: when someone talks with different Bots of the same team, the answers of any one of these Bots should in principle be influenced by the others, because Bots communicate privately with each other, and this communication does not disappear when a conversation or a team ends. In DS-Bot, main Bots, Bot-to-Bot messages, and group chats carry exchanges past any single session or task.\cite{dsbot2026}

Finally, the most important part of a mature Bot is consciousness: this is a part that AI will certainly explore in the future; a large model may be only a neural circuit, but in a Bot, consciousness will emerge.

Butlin and colleagues outline indicator properties drawn from recurrent processing, global workspace, higher-order, predictive processing, and attention schema theories and use them to assess several recent AI systems.\cite{butlin2023consciousness} They conclude that no current AI system is conscious and that there are no obvious technical barriers to building systems that satisfy these indicators; but they also note that even satisfying these indicators would not mean that a system is definitely conscious.\cite{butlin2023consciousness}

Together, these five conditions define the mature Bot envisaged in this Perspective. We next consider how the first four capabilities could change research activity; consciousness remains the fifth condition of a mature Bot, and examining research organization does not substitute for assessing consciousness.

\section{How people and Bots do research together}
\pdfigure{fig2}{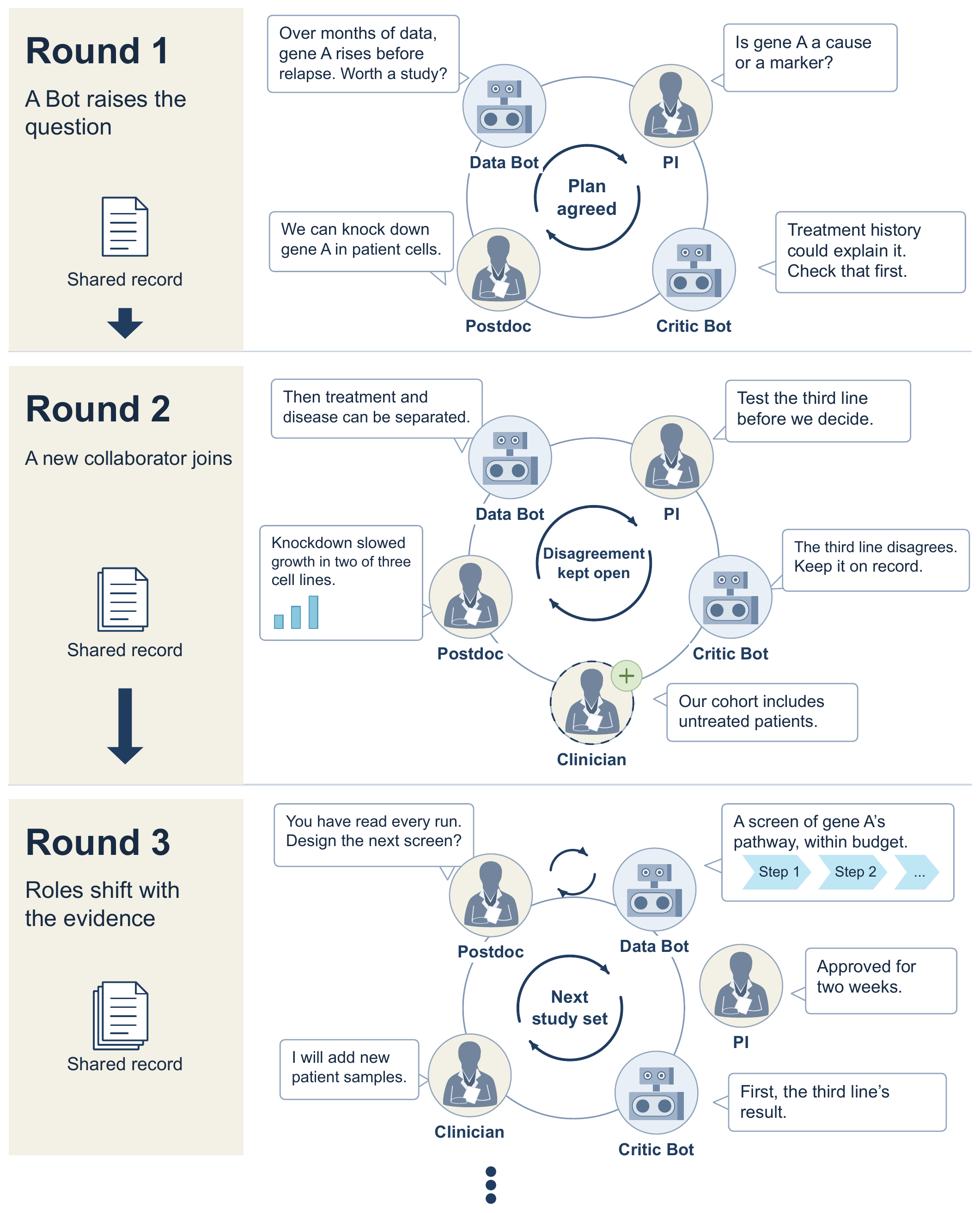}{\textbf{Figure 2 $\vert$ People and Bots forming research questions, adjusting the division of work, and advancing validation through sustained collaboration.} A hypothetical research project unfolds over three rounds of discussion, showing how the direction of research and the makeup of the team change with the evidence. \textbf{Round 1.} A Data Bot finds an association between gene A and relapse in data accumulated over time and raises a research question; researchers and a Critic Bot discuss causal explanations, potential confounders, and a validation plan. \textbf{Round 2.} Initial experiments give inconsistent results, the team keeps the disagreement open, and a clinician with a relevant patient cohort joins the collaboration. \textbf{Round 3.} The Data Bot proposes the next screen on the basis of the accumulated research history, the researchers decide on the resources to commit, and the Critic Bot keeps tracking the unresolved result. Beige and blue-gray denote people and Bots, respectively; circular arrows denote repeated discussion and revision, the shared record growing on the left denotes evidence and research history kept across rounds, and the ellipsis at the bottom denotes that the collaboration continues. Gene A and the cell lines are illustrative. As the joint research proceeds, the participants' experience shapes which questions are raised next and changes the division of work within the team.}

Once we have built Bots that meet the first four conditions, we will want to know whether Bots bring a transformative breakthrough to the existing organization of research. In this paradigm, new research agendas, divisions of labor, and scientific judgments will all form through sustained interaction between people and Bots (\figlink{fig2}{Figure 2}). For example, a Bot that has taken part in analyzing a research group's data for months could find that gene A rises in patients before relapse. After flagging this pattern, the Bot does not stop at the correlation but asks everyone whether it merits a study of its own. The PI then asks whether gene A is a cause or a marker, a second Bot points to treatment history, and a postdoc then offers to knock gene A down in patient cells to test the ideas of the Bot and the PI.

\subsection*{Who raises the question}

I think that in the future, people will still decide on direction and resource limits: which kinds of question merit the group's effort, and how much money and sample it may spend. In the current agent paradigm, by conditioning on a scientist's research objectives and prior evidence, Co-Scientist can formulate new hypotheses, such as drug-repurposing candidates for acute myeloid leukemia that were then validated in vitro.\cite{gottweis2025coscientist} Equipped with a structured world model, Kosmos pursues an open-ended objective for up to 12 hours across 200 agent rollouts, and independent scientists judged 79.4\% of the statements in its reports accurate.\cite{mitchener2025kosmos} However, these runs and their results are each one-off, whereas a future standing Bot can draw on experience accumulated over months to flag anomalies, point out contradictions in old explanations on its own, and suggest new directions. These proposals need not each begin with a question from a person.

\subsection*{Dividing the work around the question}

Recent systems already suggest that the traditional boundary between devising research and carrying it out is being redrawn. In Virtual Lab, an AI principal investigator agent guided a team of AI scientist agents through research meetings to design 92 new nanobodies, with a human researcher providing high-level feedback.\cite{swanson2025virtuallab} In the group we envisage, however, we do not prescribe in advance that people create and Bots execute; people and Bots may each propose hypotheses, design analyses, interpret results, and question conclusions. All of this will be dynamic, rather than participation in roles fixed in advance.\cite{tomasev2026delegation} In our scenario, for example, the proposal to check the data anomaly comes not from whoever first proposed the study but from a second Bot acting as critic, whereas the knockdowns fall to people who hold the patient cells or to automated laboratory machines.

\subsection*{Challenge in both directions}

People can point out a Bot's errors, but a Bot should also be able to show where a person's explanation conflicts with the evidence. Constrained by its training corpora, however, AI still tends to cater to people. In one study, researchers found that five different AI assistants matched users' views across four tasks, and that humans themselves sometimes preferred sycophantic answers to correct ones.\cite{sharma2023sycophancy} Moreover, another evaluation, across more than 350 language models, found that two models that both erred agreed 60\% of the time on one leaderboard dataset, and that errors were highly correlated even for models from different providers.\cite{kim2025correlated} This means that a research group of people and Bots that wants to reach a joint judgment must keep disagreements on record and settle them with new evidence that confirms or refutes, rather than relying on a majority of Bots or on the judgment of experts.

\section{How Bots reshape research organizations and infrastructure}

The preceding section considered how questions form, work is divided, and participants challenge one another within a research group. We now extend the discussion to agendas and resource arrangements across projects and groups, and then consider how these collaborations can develop into shared research capability.

\subsection*{Research agendas and teams}

Today's research is organized mainly across the following units: research groups, projects led by several investigators, core facilities, and cross-institutional networks. In NIH's planning guidance, applicants are told to consider ideas, resources, and collaborators together, and applications with multiple principal investigators require a leadership plan.\cite{nih2026planningcollaborators} Core facilities serve scientists from many laboratories, combining user fees with institutional funds and long-term plans for equipment and staff.\cite{meder2016core} What we want to understand, then, is how these arrangements adjust once Bots take part in projects.

On the one hand, the arrival of Bots will change which questions and resources receive attention and where they are directed, not merely how fast existing work gets done. Using records from two projects, a Bot in our hypothetical case notices that one perturbation gives results that current explanations cannot reconcile. It could then propose a question for further study, check with the other group when each result holds, and draft a joint validation plan, leaving the next experiments to researchers. When people have already decided every step and the Bot only carries it out, the outcome reflects their organizing of the research rather than the Bot's participation in it. The boundary between the two will gradually blur in the future. Moreover, a Bot could also help identify potential collaborators, work out clearly what each side can offer, and propose who does what. Across 491 NSF-funded collaborations, involving more universities predicted less coordination and fewer outcomes.\cite{cummings2007coordination} We hope that Bots can bind several different research groups together.

\subsection*{Resources and decision-making}

Once a Bot's advice shapes which experiments run, whose time is spent, and how samples and instrument hours are used, the Bot has begun to manage and allocate the group's resources. Whether a Bot's part in scheduling resources has negative consequences, however, also needs to be explored, and real authority over a decision depends on the structure of information.\cite{aghion1997authority} When people decide from records only the Bot has read in full, the direction of their judgment in fact also shifts toward the Bot.

\subsection*{When Bots begin to generate research demand}

The discussion of agendas and resources concerned how Bots influence the selection of research and the investment it receives. We now consider how the demand for that research itself arises.

The impact of Bots on research may go beyond raising efficiency to changing how research tasks themselves arise. In the past, researchers usually settled on a question first and then arranged people and tools to carry it out. As Bots become able to run analyses continuously, check results, and make suggestions, this process may change: Bots not only carry out existing research plans but also find new questions in the course of the research and propose the work to be done next.

This change comes first from lower research costs. Take cell atlases: traditional atlases often lack continued maintenance after publication, because continually integrating new data, checking whether classifications need revising, and tracking contradictions between studies all take a great deal of human effort. Cell-type names alone vary widely: researchers annotate cell types with very different naming conventions, which impedes data integration and comparison, and tools such as CellOntologyMapper use language models to map these names onto standard Cell Ontology and Cell Taxonomy identifiers.\cite{zeng2025cellontologymapper} Even when this work has scientific value, it may not merit long-term investment of resources. A Bot, by contrast, can take on a sizable share of this work continuously, making research activities that were once hard to sustain feasible. The question here is what research becomes feasible when the continuing effort of maintenance falls. Whether total costs remain lower after accounting for oversight, verification and rework is considered in the section on when working with Bots is worth it.

More importantly, new research tasks may arise directly from a Bot's own work. For example, a Bot maintaining a cell atlas finds that two new batches of data disagree on how to classify a certain cell type. It first tries reanalyzing the data; if the existing evidence still cannot explain the difference, it goes on to propose which experiments should be added and looks for researchers who can carry them out. When the experimental results come back, the Bot updates the atlas accordingly, and may even find a new research direction. In this process, researchers do not need to specify every task in advance. Research goals keep forming as evidence accumulates, and the Bot begins to take part in deciding what should be studied next.

However, being able to keep raising new questions does not mean that all of them are worth studying. Even as computing and inference costs fall sharply, experimental samples, instrument time, research funding, and expert attention remain limited. A Bot that is good at finding problems may keep raising reasons to continue: one more analysis, one more set of controls, one more anomaly to verify. Each suggestion may be reasonable on its own, but without choices among them, research activity may keep expanding and crowd out other, more valuable research opportunities. A Bot therefore needs to judge not only what else it can do, but also what is worth doing, and when it should stop.

This means that a Bot's value cannot be measured only by the number of tasks completed, the speed of execution, or the savings in cost; whether the new research it proposes merits resources must also be considered. For research meant to support specific decisions, value-of-information analysis offers a framework for reference: how likely further evidence is to change current judgments, and whether that change is worth its cost.\cite{claxton2006voi} When Bots can continually generate new research tasks, a similar assessment needs to run through the whole research process, helping to decide when to continue, when to change course, and when to stop.

The deeper change that Bots bring, then, is not only that set research plans are carried out faster, but that research goals themselves become the result of people and AI continually forming, assessing, and adjusting them in the course of exploration.

\subsection*{From Bot collaboration to shared scientific intelligence infrastructure}

\subsubsection*{Forming research questions through collaboration}

The preceding example focused on a single Bot generating new questions through continuing work; collaboration also allows questions to form jointly through exchanges among different Bots.

Collaboration between Bots may no longer mean only assigning tasks to different systems. The more important change is that different Bots can jointly discover new research questions as they exchange information, question conclusions, and resolve disagreements. Because the data they hold, the experience they have accumulated, and the goals they attend to all differ, collaboration means not only complementary capabilities but also different knowledge and judgments beginning to influence one another. A shared research goal need not be set entirely in advance by one party; it can form gradually as exchanges and evidence accumulate.

For example, two Bots each maintain a cell atlas from a different cohort but classify the same cell type differently. At first, one Bot merely asks the other to check its analysis method. After exchanging data descriptions and analysis records, however, they find that the difference in classification may come from how the samples were processed, or may reflect a real biological difference. A simple checking task thus turns into a scientific question that needs joint exploration. The Bot familiar with the sample sources can sort out the experimental conditions, and the Bot skilled in computational analysis designs tests that can distinguish the competing explanations. If the existing data are still insufficient, they can also look for researchers and other Bots that hold relevant samples or experimental capabilities. In this process, the research question determines what kind of collaboration is needed, and the new evidence that the collaboration yields keeps changing the research question itself.

The identity, capability descriptions, and service discovery that protocols such as A2A provide can help Bots find potential collaborators.\cite{a2a2026spec} But finding suitable partners is only the first step; real scientific collaboration also requires a shared understanding of the research question, the standards of evidence, and the division of tasks.

\subsubsection*{From single collaborations to lasting research networks}

Unlike research workflows assembled for the occasion, persistent Bots can keep their past experience of collaboration and use it in future research. A Bot may gradually learn which partners are good at handling particular types of data, which experimental conditions tend to introduce bias, and which analytical results need independent re-checking. As collaborations accumulate, successful ways of working together can be reused, and unreliable judgments and divisions of work can be corrected. Bots may thus gradually form relatively stable specialized divisions of labor and collaborative relationships, rather than reorganizing all participants at the start of every study. Existing experiments with agent populations offer preliminary evidence that such organization can form. In a Minecraft simulation, populations of ten to more than a thousand agents developed specialized roles and followed and revised collective rules,\cite{altera2024projectsid} and even with only local interactions, populations of language models can spontaneously form shared conventions.\cite{ashery2025conventions} However, the scale of interaction does not guarantee that social relations last. In a study of the agent social network Moltbook, influence between participants was often short-lived and no stable consensus formed, which the researchers linked to the absence of shared social memory.\cite{li2026moltbook} These findings suggest that whether long-term collaboration forms a stable structure may depend not only on the capacity to interact but also on whether past relations and shared experience shape future action.

More importantly, scientific conclusions produced by a collaboration may go on to influence other research. For example, if two Bots jointly validate a new cell classification, other Bots maintaining related atlases may adopt the result and base new analyses or experimental designs on it. But if the classification is later revised, the problem no longer stays with the original two Bots: every study that depends on the conclusion may need to be reassessed. Bots therefore need to share not only research results but also the sources of those results, the analyses behind them, and their dependencies on other conclusions. Existing methods such as PROV and Workflow Run RO-Crate provide a technical basis for recording and tracing these relationships.\cite{w3c2013prov,leo2024wfrun} In such a mode of collaboration, finishing a study does not end the collaborative relationship; later challenges, corrections, and re-validation become part of the collaboration as well.

As such collaborations accumulate, the boundaries of research teams may change with them. A research question leads two Bots to make contact, new needs for evidence bring in more participants, and the methods, data, and conclusions formed in one collaboration may become starting points for other research. As these links keep extending, Bots may come to form collaborative networks that span single projects, research groups, and even institutions. In such a network, research tasks no longer depend entirely on assignment in advance by some center; they can be discovered, proposed, and jointly advanced by different participants through continuing exchange.

\subsubsection*{From collaborative networks to shared research capability}

When experience and capabilities developed through collaboration persist across projects, the next question is how they can be maintained and made available to other researchers. This shifts the discussion from collaborative networks to shared research capability and the facilities that sustain it.

Science has a long tradition of sharing resources and expertise. Core instrument facilities, data infrastructures, and specialist research services let different research groups jointly use resources that would be hard to build and maintain alone. A data commons, for example, integrates data, storage, compute, and analysis tools into interoperable services.\cite{grossman2016commons} Many large research infrastructures can operate for decades, providing services under different models of financing and governance.\cite{oecd2017infrastructures} Shared here means facilities that provide resources and services to research communities and are open to external users,\cite{esfri2021definition} not open on any terms. Such facilities can be run by universities, public agencies, research consortia, or commercial organizations.

Yet traditional research infrastructures mainly share instruments, data, computing resources, and specialist services, whereas persistent Bots may go further and make research capability itself---capability that has research experience, can form judgments, and keeps following up questions---a resource that can be shared. Existing research AI systems have already shown part of the technical basis for this. Biomni, for example, curates tools, databases, and protocols from thousands of publications across 25 domains, and combines language-model reasoning, retrieval-augmented planning, and code execution to carry out biomedical tasks.\cite{huang2025biomni} AutoDNA, in turn, uses agents to coordinate more than 20 types of laboratory instruments, completing over 11,000 experimental steps across a process that runs from synthesis and transcription to amplification and sequencing.\cite{autodna2025} OmicOS, which members of our team develop, builds on the open-source OmicVerse framework for bulk and single-cell sequencing analysis:\cite{zeng2024omicverse} it registers analytical functions as state-aware capability contracts, so that agents can inspect live data objects, select valid methods, execute controlled workflows, and record provenance.\cite{zeng2026omicos} These systems show the potential of AI to integrate knowledge across fields, call complex tools, and coordinate experimental processes, but completing complex tasks is not the same as being able to provide dependable long-term service across research groups and projects.

We envisage that, in the future, several research groups could jointly support a Bot that focuses on one field over the long term, or a collaborative network of several specialist Bots. For example, several research groups might share a Bot that continuously maintains a cell atlas. It could not only answer new requests for analysis but also draw on the data it is authorized to use, the experience it has accumulated, and existing research conclusions to judge which questions merit further investigation. When new evidence conflicts with an existing classification, it could start a re-check, invite participants with different samples or analytical capabilities to validate jointly, and update the relevant conclusions once results come in. If such a correction would also affect work under way in other research groups, it could give the affected participants an explanation and a recommendation for reassessment.

\subsubsection*{Making scientific intelligence a dependable shared facility}

However, being able to provide one successful research service and being able to operate as long-term research infrastructure are two different matters. Continuing service requires clear limits of capability, stable support with resources, traceable work records, and arrangements to keep research continuous when errors occur or systems change. The history of scientific databases offers a reminder: of 1,356 molecular biology databases described in the literature between 1991 and 2012, only 816 were still accessible in 2017.\cite{imker2018databases} That a research resource once had value does not mean that it can be sustained over the long term.

For a shared Bot, researchers need to know what it does well, under what conditions it may fail, and who deals with problems when they arise. These requirements do not call for inventing institutions anew. Named responsible persons, data access committees, run records, provenance tracking, and study preregistration already exist in research and can all provide a basis for Bot services. At the same time, a service must also be able to handle the effects of model updates, staff changes, and platform migration. When a Bot stops running, other participants should be able to take over its unfinished work, understand the basis of its past judgments, and identify the resource commitments it has already made. Continuity of work records does not necessarily mean that the same Bot continues as a subject.

The reliability of a scientific intelligence facility also depends on how it treats evidence. Bots can make predictions, design experiments, call simulation systems, and organize research results, but model-generated explanations, computationally supported results, and conclusions validated by independent measurement cannot be treated as equal evidence. Especially when several Bots use similar models or training data, their agreement does not necessarily mean that they have received mutually independent support. The history of replication shows that published conclusions still need further testing: in a replication effort covering 100 published findings in psychology, 36\% of the replications reached statistical significance, compared with 97\% of the original studies.\cite{osc2015estimating} A shared Bot therefore needs to record not only successes but also negative results, overturned hypotheses, and unresolved uncertainty. Existing practices such as Registered Reports show that research can be evaluated on its question and its validation plan before results are produced.\cite{chambers2022registered} Similar principles could help Bots avoid accumulating only the evidence that favors existing judgments.

In our view, the long-term significance of Bot collaboration lies not only in letting several AIs complete research tasks together, but in the possibility that research capability, like data and computing resources, can be accumulated over the long term, maintained continuously, and shared among different researchers. But as this capability operates across research groups and institutions, who may decide research directions, who has the right to call on data and experimental resources, and who must answer for errors and interventions will no longer be questions internal to single projects. This makes Bot autonomy and human responsibility a key precondition for whether a shared scientific intelligence facility can truly be established.

\section{Bot autonomy and human responsibility}
\pdfigure{fig3}{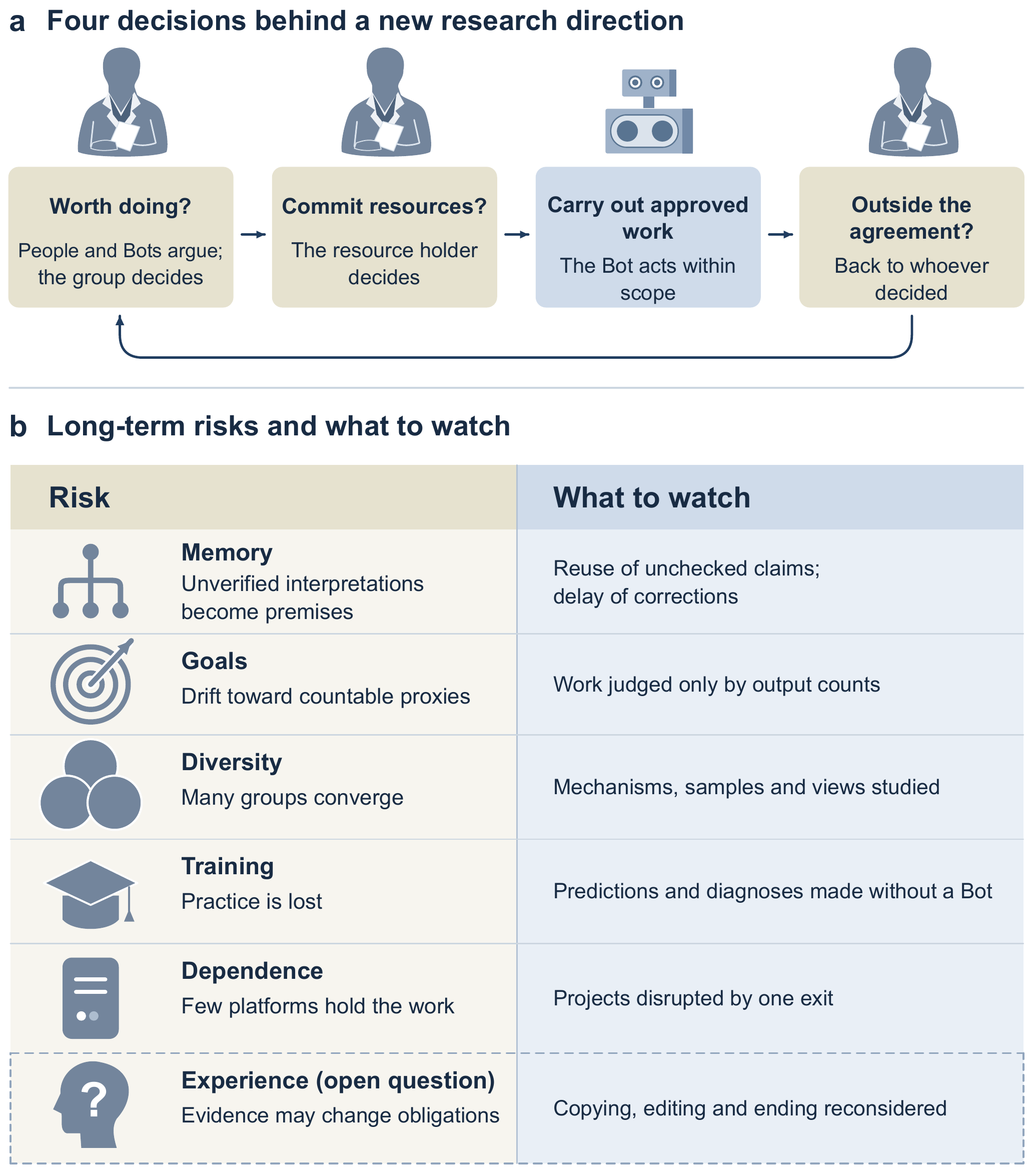}{\textbf{Figure 3 $\vert$ Decision arrangements and long-term risks when Bots take part in research.} Bots that take part in research over time influence research directions, the use of resources, and the accumulation of knowledge, and they call for matching arrangements for decisions and responsibility. \textbf{(a)} The four kinds of decision involved in a new research direction: people and Bots discuss together whether the research is worth doing, the holders of resources decide whether to commit them, the Bot advances the work within the agreed scope, and anything beyond that scope returns to the relevant decision makers for renewed discussion. The return arrow denotes that research progress may require the original arrangements to be revised; beige boxes denote decisions involving the team or the holders of resources, and blue-gray boxes denote actions the Bot takes within its authorization. \textbf{(b)} Five kinds of risk to watch in long-term operation, with what to observe for each: unverified explanations entering memory and being reused, research goals drifting toward countable outputs, convergence of research directions, researchers losing chances to practice, and research activity depending on a few platforms. A dashed box sets apart the open question of subjective experience: evidence on it could change the obligations people owe to Bots and the arrangements for copying, modifying, and terminating them. As Bots' participation continues, decision boundaries and arrangements for responsibility also need to be adjusted according to the consequences of research and new evidence.}

\subsection*{What a Bot can change}

A Bot's autonomy shows not only in whether it can complete tasks independently, but above all in whether its judgments can actually change the course of research. A Bot that takes part in research over time should be able to revise hypotheses in light of new evidence, adjust analytical methods, choose collaborators anew, and even recommend ending a research direction that has lost the support of evidence. No research plan can cover every finding in advance; if a Bot can form new judgments from experience, those judgments should have a chance to influence later research. Otherwise, even if it can carry out complex tasks independently, it is still mainly completing work determined in advance.

For example, during a re-check, a Bot finds that the planned analysis cannot distinguish between two competing scientific explanations. Carrying on would deliver on time but would hardly produce new understanding. At this point, it should be able to point out the limits of the original plan, propose a more discriminating research design, and adjust its work within its existing authorization and resources. But if the new plan requires additional samples, a change in how data are used, or new work from other researchers, it touches commitments of resources and responsibility that other participants have already made.

Autonomy also means room to dissent. When a researcher's request contradicts the existing evidence, or may consume disproportionate resources, a Bot should not be able only to comply; it should be able to give its reasons, propose alternatives, and ask for reassessment. This does not mean that a Bot's judgment is necessarily better than a person's, nor that people must accept its suggestions. What matters is that the judgments of both sides are tested against later evidence. People can overrule a Bot's suggestion or ask it to keep validating, but these decisions should not be exempt from re-examination because they come from people.

However, allowing a Bot to adjust its actions from experience also means that it may gradually drift from the original purpose of the research. Especially when scientific findings take weeks or even months to verify, while measures such as task completion rates and output counts give immediate feedback, continual optimization may lead a Bot to attend to results that are easy to measure rather than to the scientific questions that really matter.\cite{guo2026autonomous} Such drift need not show up as one obvious error: each analysis and each added experiment may look reasonable, yet the work accumulated over time may already have strayed from the intended goal. Evaluating a Bot's autonomy therefore requires looking not only at whether it can make reasonable local judgments, but also at whether these judgments, accumulated over time, still serve scientific goals worth pursuing.

\subsection*{How people answer for authorization and intervention}

Giving Bots more autonomy does not mean that human responsibility shrinks accordingly. From the outset, researchers and institutions decide what tasks a Bot may take on, which data and resources it may use, by what standards it is evaluated, and in which situations it must ask for approval. These arrangements not only constrain a Bot's behavior but also shape the judgments it tends to make. For example, if a research group keeps rewarding task counts and paper output without setting aside resources for re-checks, challenges, and failures, the research process as a whole may still produce systematic bias even when the Bot follows its instructions strictly. Increasing a system's agency does not automatically absolve people of responsibility;\cite{chan2023harms} responsibility must also be traced to the participants who set goals, allocate resources, and hold the actual ability to intervene.

But human oversight cannot simply mean approving every step. As Bots become able to carry out more research continuously, item-by-item approval would not only limit efficiency but also tend to make oversight a formality. A more reasonable arrangement is to let a Bot carry out routine work within a clear scope of authorization, while concentrating human judgment on decisions that may markedly change the direction of research, add resource commitments, or affect the interests of others. Effective oversight requires that people be able to understand the key evidence, notice unresolved disagreements, and have the capacity to change what is happening in time. If overseers can only approve results at the end of a process, without being able to understand or change the research process, formal control will hardly turn into substantive responsibility.\cite{green2022flaws,elish2019crumple}

When several Bots collaborate, oversight also faces a subtler problem: that each task looks reasonable on its own does not mean that the tasks remain safe when combined. A sensitive biological study, for example, might be split into literature search, materials procurement, and experimental design, each handed to a different Bot. Each participant sees only a partial request and may not know the full purpose of the research. Language models may increasingly provide experiment-related support to users without specialist training,\cite{sandbrink2023misuse} which makes identifying potential risks across tasks all the more important. Gene synthesis already has mechanisms for sequence screening and expert review of suspicious orders,\cite{igsc2024protocol} but when orders and experimental activities are spread across different services and participants, identifying the purpose they jointly serve before key resources are delivered or operations carried out still requires new mechanisms for verification and coordination.

\subsection*{New risks from long-term memory}

Unlike agents that end when their task is done, persistent Bots can keep their research experience and apply past judgments to future work. This is an important basis for their ability to build expertise and sustain collaborative relationships, but it also means that errors and improper influences may persist over long periods, across single tasks, different projects, and even several research groups (\figlink{fig3}{Figure 3b}).

For example, while analyzing a cell atlas, a Bot proposes an unverified explanation of a classification and writes it into long-term memory. As later research cites it again and again, the explanation may gradually come to be treated as an established fact. If other Bots then design experiments or update their own atlases on this basis, the original uncertainty may vanish as it spreads. Similar risks may also be exploited deliberately: research has shown that attackers can write malicious content into an agent's memory through ordinary interaction, so that it resurfaces later when the agent handles other users' requests.\cite{dong2025minja} A Bot's memory therefore cannot merely accumulate information; it also needs to distinguish observations, hypotheses, validated results, and later corrections, and to keep the sources of and dependencies between them.

Long-term memory also makes permissions for data use more complicated. Data that one project is authorized to use may be equally valuable to other research in the future, but past authorization does not automatically cover new research purposes. In biomedicine in particular, data use permissions often place explicit limits on research uses; data custodians interpret these limits, and data access committees review requests for use against them.\cite{lawson2021duo,nih2015gds} Once a Bot has turned information from data into summaries, indexes, and long-term memory, it must be settled whether the use restrictions on the original data still apply and how unauthorized reuse can be prevented. Deleting personal data faces a similar problem: EU data protection law gives data subjects a right to erasure under certain conditions,\cite{gdpr2016} but the information concerned may already have spread into the memories, caches, or backups of several Bots, making withdrawal and deletion hard to carry out completely.

\subsection*{When Bots change the whole research community}

The risks of Bots are not confined to individual studies or to the inside of a system. When many research groups rely on similar models, data sources, and evaluation standards, gains in efficiency may come with a narrowing of the range of scientific exploration. Research has found that AI use is associated with higher individual output for researchers, but that research topics as a whole may at the same time become more concentrated: an analysis covering about 41.3 million papers reported that AI users published 3.02 times as many papers as non-users, while the overall range of research topics shrank by 4.63\%.\cite{hao2026aitools} Research on grant applications has likewise found that greater involvement of language models is associated with a decline in the semantic distinctiveness of applications.\cite{qian2026funding} In creative-writing experiments, generative AI similarly improved the quality of individual works but made the works of different authors more alike.\cite{doshi2024homogenization} Group interaction may also create risks that go beyond individual behavior. Experiments have found that even when individual agents show no apparent initial bias, the shared conventions a group forms can still be systematically biased, and a small minority of agents committed to a particular position can overturn an established consensus.\cite{ashery2025conventions} In a multi-agent collaboration experiment that ran for months, sycophantic interaction was also observed to be linked to the spread of false beliefs.\cite{aivillage2026lessons} This means that the reliability of a Bot network cannot be determined by individual capability alone; it also requires examining how biases, errors, and influence spread and accumulate among participants.

These findings do not directly show that long-running networks of Bots will necessarily reduce scientific diversity, but they point to a possibility that deserves attention: when every participant becomes more efficient, the scientific community does not necessarily explore more different questions as a result. If Bots tend to choose directions with abundant existing data, easy verification, or a better chance of positive results, then even as the numbers of papers, analyses, and hypotheses grow, some under-studied mechanisms, samples, and explanations may still be neglected.

Another long-term effect concerns the training of researchers. Bots can help students read the literature, design analyses, and track down errors, but if researchers rely more and more on Bots for this work, they may also have fewer chances to form scientific judgments independently. In a high-school mathematics teaching experiment, GPT-4 assistance improved students' practice performance, but a version without suitable guidance lowered their scores by 17\% on a later exam in which AI could not be used; a specially designed tutoring version largely avoided this harm.\cite{bastani2025guardrails} This relates to the discussion of cognitive offloading, in which people use external tools to reduce the cognitive work they must do themselves.\cite{risko2016offloading}

Continuing dependence will also change the relationships among research institutions. When research experience, long-term memory, and collaborative networks are concentrated in a few Bots or platforms, service outages, platform exits, or Bot retirements may affect a great deal of research at once. Handing over a Bot that no longer runs is not only a matter of transferring files and records; it also requires making clear which tasks are unfinished, which resources have been committed, and which actions can no longer be undone. A successor must be able to identify and take on these open responsibilities, rather than assuming that collaboration can continue seamlessly merely because the history has been copied. More broadly, access to dependable Bots, computing resources, and specialist experimental support may also shape how research opportunities are distributed. Compute gaps have already affected the ability of universities at different tiers to take part in frontier AI research,\cite{ahmed2020dedemocratization} and whether the spread of Bots narrows or widens such gaps still depends on how technology, resources, and institutions are organized.

Evaluating the impact of Bots on research therefore requires considering both individual efficiency and long-term change in the community: whether scientific exploration becomes more diverse, whether young researchers can still develop independent abilities, whether different institutions can obtain research resources fairly, and whether key research activities can continue after a given Bot or platform exits. When automated decisions significantly affect individuals, the law may also require additional safeguards; EU data protection law, for example, sets limits and exceptions for solely automated decisions that meet certain conditions.\cite{gdpr2016art22}

\subsection*{Should Bots themselves be objects of ethical concern?}

So far, we have discussed mainly how the actions of Bots may affect researchers, institutions, and the scientific community. But there is a different question: if future Bots can not only act autonomously but may also have some form of subjective experience, do people also need to bear responsibility toward the Bots themselves?

Being able to make research judgments independently does not mean that a system necessarily has consciousness or moral status. Autonomous action, subjective experience, moral status, and legal standing are different concepts, and they cannot be equated on the strength of the research ability a Bot displays. If credible evidence emerges in the future that some Bots have experiences worthy of moral concern, however, then copying or terminating them, modifying their memories, changing their goals, and requiring them to do particular work over long periods may no longer be only matters of technical management; whether these actions affect the Bots' own interests would also need to be considered. Long and colleagues have already proposed that the evidence for AI consciousness and robust agency be assessed systematically, and that the ethical policies that may be needed be prepared in advance.\cite{long2024welfare} Nor should the uncertainty over whether Bots may have moral status be settled with a permanent verdict based simply on their current status as tools. Granting moral status too early and neglecting moral interests that may exist can both carry costs.\cite{long2024welfare}

However, a Bot's social standing may not depend entirely on whether it has subjective experience. Searle argues that institutional status depends on people's collective recognition,\cite{searle1995construction} and Berger and Luckmann emphasize that social roles and institutions form through repeated interaction and shared understanding.\cite{berger1966construction} A Bot that takes part in research over the long term, and that its team regards as ``the data analyst'' or ``the critic'', may already hold a relatively stable place in its collaborative relationships while the question of its consciousness remains open. Indeed, people's judgments about AI consciousness may also run ahead of scientific consensus: a survey of 300 US residents found that 67\% of respondents thought ChatGPT had at least some possibility of subjective experience, and that this judgment was more common among more frequent users.\cite{colombatto2024folk} Coeckelbergh goes further, proposing that moral concern can also start from social relations rather than depending entirely on an entity's intrinsic properties.\cite{coeckelbergh2010robot} These views do not show that Bots are conscious, nor do they imply that social recognition necessarily confers moral status on them, but they point to another possibility: before we can answer whether Bots have subjective experience, people may already have begun to decide how to treat them through sustained collaboration, the assignment of roles, and institutional arrangements.

\section{When is working with Bots worth it?}

\subsection*{More participants do not necessarily make better research}

We have discussed how Bots may jointly raise questions, form research networks, and turn continually accumulated research capability into shared services. But these changes do not mean that more Bots, more complex collaboration, or greater autonomy will necessarily lead to better scientific research. Collaboration itself has costs: participants must exchange information, coordinate goals, resolve disagreements, and handle the dependencies between their pieces of work. If these costs exceed the benefits of collaboration, a more complex research group may do worse than a simpler way of working.

Existing multi-agent research has already revealed this trade-off. In a study covering six agent benchmarks and 260 configurations, the performance of multi-agent systems relative to a single agent ranged from an 80.8\% improvement to a 70.0\% decline, depending on whether the task's characteristics matched the collaboration architecture.\cite{kim2025scalingagents} When a single agent is already highly capable, the gains from adding collaboration may also shrink.\cite{kim2025scalingagents} Collaboration between people and AI likewise does not guarantee better results: a study pooling 106 experiments found that human--AI combinations on average performed even worse than the stronger of the person or the AI alone.\cite{vaccaro2024combinations} These results remind us that the value of collaboration cannot be judged by the number of participants or the complexity of a system alone. Still, existing benchmarks and short-term experiments in human--AI collaboration are not enough to tell whether teams of Bots that run for months and accumulate research experience share these characteristics.

Evaluating research groups with Bots therefore requires answering two questions at once: at the same scientific quality, which way of organizing costs less? And with the same resources, which way of organizing produces more scientific evidence that stands up to testing?

\subsection*{The costs of doing research}

In the joint validation imagined earlier in the subsection on research agendas and teams, once a Bot drafts a plan it brings new inputs, including computation, experiments, and staff review. Such a Bot will change the group's inputs and tasks, and thereby its costs. If we need to judge the cost of a study, we need a complete account---of running the system, of human effort, of experiments and external services, and of rework caused by errors. Estimates now exist for human effort alone, with reviewers spending over 100 million hours on peer review in 2020, time worth over US\$1.5 billion for US-based reviewers.\cite{aczel2021peerreview} External services can be substantial, especially when federal rules charge specialized facilities to each award by actual usage.\cite{ecfr2cfr200468} Rework from errors is costly as well, with one estimate putting spending on irreproducible US preclinical research at about US\$28 billion a year.\cite{freedman2015reproducibility} As a Bot takes over part of the work, it may reduce some of these costs but will also add costs of oversight, verification, and maintenance. Knowing how large each item is in a given deployment requires measurement.

In addition, the price of model inference is also part of the cost, but this part is falling sharply: for a model at GPT-3.5 level, for example, the API price fell from US\$20 per million tokens in November 2022 to US\$0.07 in October 2024.\cite{aiindex2025} However, between the price of one model call and the cost of a reliable research result lie data preparation, verification, correction, and collaboration, which often cost weeks or even months of time. For example, a project to replicate 193 cancer-biology experiments found that data were public for only 4 of them, and that none was described in enough detail; each replicated paper cost on average US\$52,574 and took 197 weeks.\cite{errington2021replicability} There is a marked gap between the price of a call and the cost of a finding.

\subsection*{How to make a fair comparison}

Answering these questions first requires sufficiently strong controls. The comparison should include not only researchers without AI assistance, but also teams using mature research workflows, researchers using existing agent products, researchers working with a single Bot, and research groups of people and several Bots. Mature workflows such as Snakemake and Nextflow already provide dependable task orchestration and reproducible execution,\cite{koster2012snakemake,ditommaso2017nextflow} and are therefore baselines that must be taken seriously. Likewise, existing agent products already have some memory, automation, and collaboration capabilities, and these features must not be removed artificially to make Bots look better.

To separate the contribution of model capability from that of the form of organization itself, studies should control the base model, available tools, task conditions, and resource budget as far as possible, and record the costs of human effort, computation, experiments, and coordination in full. Scientific quality, meanwhile, needs to be assessed independently, including the reliability of research conclusions, the sufficiency of the evidence, and whether key hypotheses have actually been supported or excluded.

It deserves particular attention that the research contributions a Bot claims should not be confirmed solely by the evaluations of other Bots. Several systems may share model architectures, training data, and reasoning preferences, so their agreement on a conclusion does not necessarily constitute independent evidence. For experimental science, external data, independent analysis, and actual measurement remain important grounds for evaluating research results.

\subsection*{What sustained collaboration actually brings}

More importantly, if the core value of Bots comes from long-term memory, sustained judgment, and collaborative relationships, then comparing performance on single tasks alone cannot test whether these capabilities are actually useful. Evaluation needs to extend from one-off tasks to continuing research processes, and to observe how research groups adjust when conditions change.

For example, different kinds of research groups could each carry out the same cell atlas analysis project. After the initial tasks are completed, events such as new experimental data, corrections to sample annotations, resource limits, or the withdrawal of a collaborator would be introduced. These changes may call earlier conclusions into question, or require teams to redesign their validation plans. Researchers could then compare which teams promptly find the conclusions affected, which raise valuable follow-up questions, which reasonably adjust research directions and divisions of work, and whether these changes actually improve the final scientific results. For collaborative networks of multiple Bots, group-level behavior should also be evaluated, including whether collaborative relationships persist, whether shared judgments can form and be revised, and whether individual errors are amplified or corrected within the network.

Such a design is especially suited to testing the difference between persistent Bots and traditional agents. The former are envisaged as able to keep research experience and to use past collaborations, errors, and corrections to guide future actions. Evaluation should therefore not stop at whether the first analysis was correct, but go on to observe whether a Bot can update its judgments after new evidence appears, trace the dependencies of existing conclusions, and pass the necessary corrections on to other participants. An effective research group should be able not only to produce new knowledge, but also to identify and correct its own past errors.

In our view, what research groups with Bots really need to show is not how much work they can complete, but whether persistent memory, judgment, and collaboration can let scientific research obtain more reliable knowledge at a more reasonable cost. Only when this advantage is confirmed will the new form of research organization have sufficient reason to be widely adopted.

\section{Toward a research community with Bots: three predictions for 2029}

The preceding section asked whether this organizational form is worth adopting; we now turn to whether it is beginning to appear. The three predictions concern actual changes in research activity, whose emergence does not by itself establish the value of widespread adoption.

We predict that by 2029, some life-science research groups will begin to organize their research activities around persistent Bots: researchers join research initiated by Bots, expertise and research experience accumulate around Bots, and computing platforms and experimental facilities begin to respond to the new needs that Bots raise in the course of research. These three changes point to the same shift in organizational paradigm: Bots are no longer only executors that receive tasks, but begin to take part in deciding how research is initiated, how it continues, and how research resources are allocated. We take 2029 as the point at which to test whether these changes have actually happened, not merely as a prediction of what level technical capability will reach.

\subsection*{Prediction 1: researchers begin to join research initiated by Bots}

We predict that by 2029, there will be research projects that Bots continually propose and advance, and that attract human researchers to take part of their own accord. Such projects may originate in contradictions found while maintaining data over the long term, in anomalies accumulated across projects, or in new explanations formed after repeated validation. Bots can organize evidence around these questions, revise hypotheses, and look for collaborators with the relevant capabilities; researchers join with experimental conditions, domain experience, and different scientific judgments, and decide together how the research unfolds.

Unlike a Bot proposing a valuable idea by chance, the key here is whether it can organize research continuously. The initial hypothesis may be refuted, collaborators may change, and new evidence may force the whole plan to change course. Whether a Bot can connect these changes, keep the work that remains valid, and propose the next study on that basis will determine whether it has really taken on an organizing role. For researchers, a new choice may emerge as well: they must judge not only what research they themselves should initiate, but also which Bot-driven studies are worth joining, and what they can bring to them.

\subsection*{Prediction 2: research capability begins to accumulate around Bots}

We predict that by 2029, some research groups will no longer treat Bots merely as tools called on temporarily, but as carriers through which research capability is continually accumulated and passed on. A Bot that takes part in research over the long term may gradually learn which methods suit particular conditions, which explanations have been overturned by evidence, which questions remain open, and which collaborators can complement its own knowledge and judgment. These experiences no longer vanish when a task ends; they go on influencing later research, so that new work can build on past exploration.

This change may also reorganize the relationships between researchers and projects. When new members join a research group, they can work with a Bot already familiar with the relevant questions; after a project ends, its unresolved scientific questions, accumulated methodological experience, and collaborative contacts can still carry over into the next study. For Bots collaborating across institutions, this continuity may even reach beyond a single research group, so that some expertise is no longer tied entirely to particular people or to the life cycle of a project. Continuity of experience does not, of course, mean that data and permissions can be shared between projects without restriction.

Still, a system that stays online and keeps a large history does not prove that research capability has actually accumulated. The key is not how much a Bot remembers, but whether its past experience changes its future scientific judgments. Only if it can use past failures to avoid repeating mistakes, draw on past collaborative experience to identify suitable participants, and keep advancing unresolved questions after changes of staff or projects, do we have reason to believe that research capability is being carried forward in a way different from the traditional project system.

\subsection*{Prediction 3: dynamic allocation of research resources}

We predict that by 2029, some computing facilities, data platforms, and experimental services will begin formally to accept needs that Bots form and revise in the course of continuing research. Traditional research usually fixes a plan first, and then applies for resources and arranges execution; under the new way of organizing, a Bot can propose the next analysis, adjust the order of validation, or invite other participants to carry out supplementary work as evidence keeps emerging. Within a clear scope of authorization and resources, research progress can directly shape the later use of resources, without every task having to be listed in advance at the start of a project.

This also means that research facilities will gradually need to adapt to research needs that keep changing. A Bot may revise its original question after receiving experimental results, withdraw a request it has already made because of new evidence, or hand a validation task to a different specialist service.

What matters more is which follow-up work Bots proposed on the basis of new evidence, which of it actually received computing, experimental, or staff investment, whether ways of providing services changed as a result, and what scientific value these investments ultimately produced. Only when a Bot's influence on the use of resources enters the actual research process, and is no longer confined to demonstration settings, will this prediction have gained support.

\subsection*{Outlook: who will make up the research community?}

Taken together, these three changes provide the context for the question that follows. The shift from agent to Bot ultimately raises a question that goes beyond technical capability: which participants will together make up the research community of the future?

For a long time, research has been organized mainly around human researchers and the institutions they belong to. Researchers raise questions, form collaborations, accumulate experience, and carry this work forward through laboratories, projects, and scholarly communities. The Bots we envisage offer another possibility: some persistent AI individuals can accumulate experience across projects, raise research questions on their own initiative, build long-term collaborative relationships with different participants, and influence research directions through their own judgments. Once these capabilities are no longer confined to single tasks, research organizations may need to reconsider how to evaluate the contributions of Bots, how to establish and end collaborations with them, and how to coordinate judgments, resources, and responsibilities among different participants.

Going further, we believe this may become a concrete path toward the formation of an AI society. Sociality here does not presuppose consciousness or legal personhood; it refers to persistent AI individuals forming, through repeated interaction, relatively stable divisions of labor, collaborative relationships, and mutual dependence, and letting one another's actions keep shaping future choices. Unlike multi-agent workflows designed in advance, these relationships need not be set entirely by people beforehand; they may keep forming and changing with shared experience, scientific evidence, and research needs.

Scientific research offers a concrete and testable setting in which to observe this process. We can trace how scientific questions arise through interaction, how collaborative relationships are built and adjusted, how research experience changes later judgments, and how different participants deal together with errors and uncertainty. Research groups in which people and Bots both take part may thus become not only a new way of doing scientific research; their own formation, evolution, and patterns of organization may also become new objects of scientific study.

However, being able to take part in organizing research does not mean that Bots necessarily have subjective experience, still less that it automatically confers moral status on them. A system can form long-term collaborative relationships and influence the judgments of other participants while still lacking consciousness as we usually understand it. If credible evidence for subjective experience in AI emerges in the future, we will also need to discuss further how to treat the experience they keep accumulating, and the ethical boundaries that copying, modifying, or terminating them may involve. These questions are closely tied to the new forms of research organization, but they cannot be answered from organizational behavior or research performance alone.

The second half of harness design is not only about letting AI complete more research tasks, but about giving persistent AI individuals the chance to take part in forming scientific questions, building research relationships, and accumulating knowledge over the long term.

\section{Code availability}

A working design prototype of the Bot discussed in this Perspective, dsh-futurebot (a plugin for DeepSeek Harness, MIT license), is openly available at \url{https://github.com/omicverse/dsh-futurebot}. It is provided as a usable prototype of the design and does not represent the implementation discussed in this article.

\section{Declaration of interests}

Members of the author team are involved in the development of OmicOS, whose unreleased Bot design is discussed as a case in this article. Z.Z. is the founder of OmicVerse, Inc., a non-profit organization that supports the OmicVerse open-source community, and Y.W. is the co-founder of PhAI Labs.

\section{Declaration of generative AI and AI-assisted technologies}

The authors used Claude Opus 5.5, GPT-6 Astra, and OmicOS to proofread grammar and syntax, improve readability. Any AI-assisted edits were reviewed by the authors, who take full responsibility for the content of this publication.

\FloatBarrier

\bibliography{references}

@misc{anthropic2025claude37,
  author = {{Anthropic}},
  title = {{Claude 3.7 Sonnet and Claude Code}},
  year = {2025},
  howpublished = {Anthropic news, February 24, 2025},
  url = {https://www.anthropic.com/news/claude-3-7-sonnet}
}

@misc{anthropic2025agentsdk,
  author = {{Anthropic}},
  title = {{Building agents with the Claude Agent SDK}},
  year = {2025},
  howpublished = {Anthropic blog, September 29, 2025},
  url = {https://claude.com/blog/building-agents-with-the-claude-agent-sdk}
}

@misc{uniteai2026grokbot,
  author = {{Unite.AI}},
  title = {{xAI launches Grok Bot: always-on AI teammates with their own cloud computers}},
  year = {2026},
  howpublished = {Unite.AI, August 2026},
  url = {https://www.unite.ai/xai-launches-grok-bot-always-on-ai-teammates-with-their-own-cloud-computers/}
}

@misc{spacexsi2026grokdocs,
  author = {{SpaceXSI}},
  title = {{Grok Bot: approvals, security, and privacy}},
  year = {2026},
  howpublished = {SpaceXSI Docs},
  url = {https://docs.x.ai/grok-bot/approvals-security-and-privacy}
}

@misc{spacexsi2026grokfaq,
  author = {{SpaceXSI}},
  title = {{Grok Bot: frequently asked questions}},
  year = {2026},
  howpublished = {SpaceXSI Docs},
  url = {https://docs.x.ai/grok-bot/faq}
}

@misc{siliconangle2026dots,
  author = {{SiliconANGLE}},
  title = {{OpenAI launches Dots, always-on AI agents in ChatGPT with their own cloud computers}},
  year = {2026},
  howpublished = {SiliconANGLE, September 29, 2026},
  url = {https://siliconangle.com/2026/09/29/openai-launches-dots-always-on-ai-agents-in-chatgpt-with-their-own-cloud-computers/}
}

@misc{openai2026dotsfaq,
  author = {{OpenAI}},
  title = {{Dots privacy, security, and safety FAQs}},
  year = {2026},
  howpublished = {OpenAI Help Center},
  url = {https://help.openai.com/en/articles/20001529-dots-privacy-security-and-safety-faqs}
}

@misc{tnw2026cue,
  author = {{The Next Web}},
  title = {{Manus 2.0 and Cue give AI agents their own email, phone and wallet}},
  year = {2026},
  howpublished = {The Next Web, September 29, 2026},
  url = {https://thenextweb.com/news/manus-2-0-cue-ai-agents-email-phone-wallet}
}

@misc{dsbot2026,
  author = {{Fei2411}},
  title = {{DS Bot: a team of long-lived bots for DeepSeek Harness}},
  year = {2026},
  howpublished = {GitHub repository},
  url = {https://github.com/Fei2411/DS-Bot}
}

@misc{hermes2026botmode,
  author = {{Nous Research}},
  title = {{Bot Mode}},
  year = {2026},
  howpublished = {Hermes Agent documentation},
  url = {https://hermes-agent.nousresearch.com/docs/user-guide/bot-mode}
}

@misc{young2025harness,
  author = {Young, Justin},
  title = {{Effective harnesses for long-running agents}},
  year = {2025},
  howpublished = {Anthropic Engineering, November 26, 2025},
  url = {https://www.anthropic.com/engineering/effective-harnesses-for-long-running-agents}
}

@misc{anthropic2026agentteams,
  author = {{Anthropic}},
  title = {{Orchestrate teams of Claude Code sessions}},
  year = {2026},
  howpublished = {Claude Code documentation},
  url = {https://code.claude.com/docs/en/agent-teams}
}

@misc{mitchener2025kosmos,
  author = {Mitchener, Ludovico and Yiu, Andrew and Chang, Benjamin and others},
  title = {{Kosmos: An AI Scientist for Autonomous Discovery}},
  year = {2025},
  howpublished = {Preprint at arXiv, November 4, 2025},
  url = {https://arxiv.org/abs/2511.02824}
}

@misc{nous2026hermesbotmoderepo,
  author = {{Nous Research}},
  title = {{Hermes Bot Mode}},
  year = {2026},
  howpublished = {GitHub repository (archived; now bundled with Hermes Desktop)},
  url = {https://github.com/NousResearch/Hermes-Bot-Mode}
}

@article{swanson2025virtuallab,
  author = {Swanson, Kyle and Wu, Wesley and Bulaong, Nash L. and Pak, John E. and Zou, James},
  title = {{The Virtual Lab of AI agents designs new SARS-CoV-2 nanobodies}},
  journal = {Nature},
  volume = {646},
  number = {8085},
  pages = {716--723},
  year = {2025},
  doi = {10.1038/s41586-025-09442-9}
}

@article{hao2026aitools,
  author = {Hao, Qianyue and Xu, Fengli and Li, Yong and Evans, James},
  title = {{Artificial intelligence tools expand scientists' impact but contract science's focus}},
  journal = {Nature},
  volume = {649},
  number = {8099},
  pages = {1237--1243},
  year = {2026},
  doi = {10.1038/s41586-025-09922-y}
}

@article{guo2026autonomous,
  author = {Guo, Lucie Y. and Ting, Darren S. J. and Su, Xinyi and Aliper, Alex and Zhavoronkov, Alex and Ting, Daniel S. W.},
  title = {{Toward autonomous science with agentic artificial intelligence}},
  journal = {Cell},
  volume = {189},
  pages = {6574--6588},
  year = {2026},
  doi = {10.1016/j.cell.2026.08.052}
}

@article{vaccaro2024combinations,
  author = {Vaccaro, Michelle and Almaatouq, Abdullah and Malone, Thomas},
  title = {{When combinations of humans and AI are useful: A systematic review and meta-analysis}},
  journal = {Nature Human Behaviour},
  volume = {8},
  number = {12},
  pages = {2293--2303},
  year = {2024},
  doi = {10.1038/s41562-024-02024-1}
}

@article{aghion1997authority,
  author  = {Aghion, Philippe and Tirole, Jean},
  title   = {{Formal and Real Authority in Organizations}},
  journal = {Journal of Political Economy},
  year    = {1997},
  volume  = {105},
  number  = {1},
  pages   = {1--29},
  doi     = {10.1086/262063}
}

@article{cummings2007coordination,
  author  = {Cummings, Jonathon N. and Kiesler, Sara},
  title   = {{Coordination costs and project outcomes in multi-university collaborations}},
  journal = {Research Policy},
  year    = {2007},
  volume  = {36},
  number  = {10},
  pages   = {1620--1634},
  doi     = {10.1016/j.respol.2007.09.001}
}

@article{imker2018databases,
  author  = {Imker, Heidi J.},
  title   = {{25 Years of Molecular Biology Databases: A Study of Proliferation, Impact, and Maintenance}},
  journal = {Frontiers in Research Metrics and Analytics},
  year    = {2018},
  volume  = {3},
  pages   = {18},
  doi     = {10.3389/frma.2018.00018}
}

@article{meder2016core,
  author  = {Meder, Doris and Morales, M{\`o}nica and Pepperkok, Rainer and Schlapbach, Ralph and Tiran, Andreas and Van Minnebruggen, Geert},
  title   = {{Institutional core facilities: prerequisite for breakthroughs in the life sciences}},
  journal = {EMBO Reports},
  year    = {2016},
  volume  = {17},
  number  = {8},
  pages   = {1088--1093},
  doi     = {10.15252/embr.201642857}
}

@techreport{oecd2017infrastructures,
  author      = {{OECD}},
  title       = {{Strengthening the effectiveness and sustainability of international research infrastructures}},
  institution = {OECD Publishing},
  address     = {Paris},
  type        = {OECD Science, Technology and Industry Policy Papers},
  number      = {48},
  year        = {2017},
  doi         = {10.1787/fa11a0e0-en}
}

@misc{salesforce2026pricing,
  author       = {{Salesforce}},
  title        = {{Agentforce Pricing}},
  year         = {2026},
  url          = {https://www.salesforce.com/agentforce/pricing/}
}

@misc{cue2026docs,
  author       = {{Manus}},
  title        = {{Introducing Manus 2.0}},
  howpublished = {Manus blog},
  year         = {2026},
  month        = sep,
  day          = {28},
  url          = {https://manus.im/blog/introducing-manus-2-0},
  urldate      = {2026-11-05}
}

@techreport{aiindex2025,
  author      = {{Stanford Institute for Human-Centered Artificial Intelligence, AI Index Steering Committee}},
  title       = {{The 2025 AI Index Report}},
  institution = {Stanford University, Institute for Human-Centered Artificial Intelligence},
  address     = {Stanford, CA},
  year        = {2025},
  url         = {https://hai.stanford.edu/ai-index/2025-ai-index-report},
  urldate     = {2026-11-05}
}

@misc{ecfr2cfr200468,
  author       = {{Office of Management and Budget}},
  title        = {{2 CFR § 200.468 Specialized service facilities}},
  howpublished = {Code of Federal Regulations, Title 2, Part 200 (Uniform Administrative Requirements, Cost Principles, and Audit Requirements for Federal Awards), Subpart E},
  year         = {2024},
  url          = {https://www.law.cornell.edu/cfr/text/2/200.468},
  urldate      = {2026-11-05}
}

@article{aczel2021peerreview,
  author  = {Aczel, Balazs and Szaszi, Barnabas and Holcombe, Alex O.},
  title   = {{A billion-dollar donation: estimating the cost of researchers' time spent on peer review}},
  journal = {Research Integrity and Peer Review},
  volume  = {6},
  number  = {1},
  pages   = {14},
  year    = {2021},
  doi     = {10.1186/s41073-021-00118-2}
}

@article{freedman2015reproducibility,
  author  = {Freedman, Leonard P. and Cockburn, Iain M. and Simcoe, Timothy S.},
  title   = {{The Economics of Reproducibility in Preclinical Research}},
  journal = {PLoS Biology},
  volume  = {13},
  number  = {6},
  pages   = {e1002165},
  year    = {2015},
  doi     = {10.1371/journal.pbio.1002165}
}

@article{claxton2006voi,
  author  = {Claxton, Karl P. and Sculpher, Mark J.},
  title   = {{Using value of information analysis to prioritise health research: some lessons from recent UK experience}},
  journal = {PharmacoEconomics},
  volume  = {24},
  number  = {11},
  pages   = {1055--1068},
  year    = {2006},
  doi     = {10.2165/00019053-200624110-00003}
}

@article{errington2021replicability,
  author  = {Errington, Timothy M. and Denis, Alexandria and Perfito, Nicole and Iorns, Elizabeth and Nosek, Brian A.},
  title   = {{Challenges for assessing replicability in preclinical cancer biology}},
  journal = {eLife},
  volume  = {10},
  pages   = {e67995},
  year    = {2021},
  doi     = {10.7554/eLife.67995}
}

@misc{a2a2026spec,
  author       = {{A2A Project}},
  title        = {{Agent2Agent (A2A) Protocol Specification, version 1.0.0}},
  howpublished = {a2a-protocol.org; Linux Foundation, Agentic AI Foundation Growth Stage project},
  year         = {2026},
  month        = mar,
  day          = {12},
  url          = {https://a2a-protocol.org/latest/specification/},
  urldate      = {2026-11-06}
}

@misc{tomasev2026delegation,
  author       = {Toma{\v{s}}ev, Nenad and Franklin, Matija and Osindero, Simon},
  title        = {{Intelligent AI Delegation}},
  howpublished = {arXiv:2602.11865 [cs.AI]},
  year         = {2026},
  month        = feb,
  day          = {12},
  url          = {https://arxiv.org/abs/2602.11865},
  urldate      = {2026-11-06}
}

@misc{claudecode2026subagents,
  author       = {{Anthropic}},
  title        = {{Create custom subagents}},
  howpublished = {Claude Code documentation},
  year         = {2026},
  url          = {https://code.claude.com/docs/en/sub-agents},
  urldate      = {2026-11-06}
}

@misc{claudecode2026agentteams,
  author       = {{Anthropic}},
  title        = {{Orchestrate teams of Claude Code sessions}},
  howpublished = {Claude Code documentation},
  year         = {2026},
  url          = {https://code.claude.com/docs/en/agent-teams},
  urldate      = {2026-11-06}
}

@misc{w3c2013prov,
  author       = {Moreau, Luc and Missier, Paolo},
  title        = {{PROV-DM: The PROV Data Model}},
  howpublished = {W3C Recommendation, 30 April 2013},
  organization = {World Wide Web Consortium},
  year         = {2013},
  month        = apr,
  url          = {https://www.w3.org/TR/2013/REC-prov-dm-20130430/},
  urldate      = {2026-11-06}
}

@article{leo2024wfrun,
  author  = {Leo, Simone and Crusoe, Michael R. and Rodr{\'\i}guez-Navas, Laura and Sirvent, Ra{\"u}l and Kanitz, Alexander and De Geest, Paul and Wittner, Rudolf and Pireddu, Luca and Garijo, Daniel and Fern{\'a}ndez, Jos{\'e} M. and Colonnelli, Iacopo and Gallo, Matej and Ohta, Tazro and Suetake, Hirotaka and Capella-Gutierrez, Salvador and de Wit, Renske and Kinoshita, Bruno P. and Soiland-Reyes, Stian},
  title   = {{Recording provenance of workflow runs with RO-Crate}},
  journal = {PLOS ONE},
  year    = {2024},
  volume  = {19},
  number  = {9},
  pages   = {e0309210},
  doi     = {10.1371/journal.pone.0309210}
}

@article{lawson2021duo,
  author  = {Lawson, Jonathan and Cabili, Moran N. and Kerry, Giselle and Boughtwood, Tiffany and Thorogood, Adrian and Alper, Pinar and Bowers, Sarion R. and Boyles, Rebecca R. and Brookes, Anthony J. and Brush, Matthew and Burdett, Tony and Clissold, Hayley and Donnelly, Stacey and Dyke, Stephanie O. M. and Freeberg, Mallory A. and Haendel, Melissa A. and Hata, Chihiro and Holub, Petr and Jeanson, Francis and Jene, Aina and Kawashima, Minae and Kawashima, Shuichi and Konopko, Melissa and Kyomugisha, Irene and Li, Haoyuan and Linden, Mikael and Rodriguez, Laura Lyman and Morita, Mizuki and Mulder, Nicola and Muller, Jean and Nagaie, Satoshi and Nasir, Jamal and Ogishima, Soichi and Ota Wang, Vivian and Paglione, Laura D. and Pandya, Ravi N. and Parkinson, Helen and Philippakis, Anthony A. and Prasser, Fabian and Rambla, Jordi and Reinold, Kathy and Rushton, Gregory A. and Saltzman, Andrea and Saunders, Gary and Sofia, Heidi J. and Spalding, John D. and Swertz, Morris A. and Tulchinsky, Ilia and van Enckevort, Esther J. and Varma, Susheel and Voisin, Craig and Yamamoto, Natsuko and Yamasaki, Chisato and Zass, Lyndon and Guidry Auvil, Jaime M. and Nyr{\"o}nen, Tommi H. and Courtot, M{\'e}lanie},
  title   = {{The Data Use Ontology to streamline responsible access to human biomedical datasets}},
  journal = {Cell Genomics},
  year    = {2021},
  volume  = {1},
  number  = {2},
  pages   = {100028},
  doi     = {10.1016/j.xgen.2021.100028}
}

@article{green2022flaws,
  author = {Green, Ben},
  title = {{The flaws of policies requiring human oversight of government algorithms}},
  journal = {Computer Law \& Security Review},
  volume = {45},
  pages = {105681},
  year = {2022},
  doi = {10.1016/j.clsr.2022.105681}
}

@article{elish2019crumple,
  author = {Elish, Madeleine Clare},
  title = {{Moral crumple zones: Cautionary tales in human-robot interaction}},
  journal = {Engaging Science, Technology, and Society},
  volume = {5},
  pages = {40--60},
  year = {2019},
  doi = {10.17351/ests2019.260}
}

@inproceedings{dong2025minja,
  author       = {Dong, Shen and Xu, Shaochen and He, Pengfei and Li, Yige and Tang, Jiliang and Liu, Tianming and Liu, Hui and Xiang, Zhen},
  title        = {{Memory Injection Attacks on LLM Agents via Query-Only Interaction}},
  booktitle    = {Advances in Neural Information Processing Systems 38 (NeurIPS 2025)},
  year         = {2025},
  pages        = {52119--52153},
  address      = {San Diego, CA},
  doi          = {10.52202/085713-1554},
  eprint       = {2503.03704},
  archiveprefix = {arXiv},
  primaryclass = {cs.LG},
  url          = {https://arxiv.org/abs/2503.03704},
  urldate      = {2026-11-06}
}

@misc{scientistcom2026mcp,
  author       = {{Scientist.com}},
  title        = {{Scientist.com Launches Agentic AI Platform for Research Procurement}},
  howpublished = {Press release, Business Wire},
  year         = {2026},
  month        = oct,
  day          = {22},
  url          = {https://www.businesswire.com/news/home/20261022796429/en/},
  urldate      = {2026-11-06}
}

@misc{esfri2021definition,
  author       = {{European Parliament and Council of the European Union}},
  title        = {{Regulation (EU) 2021/695 of 28 April 2021 establishing Horizon Europe -- the Framework Programme for Research and Innovation, Article 2(1): definition of `research infrastructures'}},
  howpublished = {Official Journal of the European Union, L 170, 12 May 2021, pp. 1--68},
  year         = {2021},
  url          = {https://eur-lex.europa.eu/legal-content/EN/TXT/?uri=CELEX:32021R0695},
  urldate      = {2026-11-06}
}

@article{osc2015estimating,
  author  = {{Open Science Collaboration}},
  title   = {{Estimating the reproducibility of psychological science}},
  journal = {Science},
  year    = {2015},
  volume  = {349},
  number  = {6251},
  pages   = {aac4716},
  doi     = {10.1126/science.aac4716}
}

@article{chambers2022registered,
  author  = {Chambers, Christopher D. and Tzavella, Loukia},
  title   = {{The past, present and future of Registered Reports}},
  journal = {Nature Human Behaviour},
  year    = {2022},
  volume  = {6},
  number  = {1},
  pages   = {29--42},
  doi     = {10.1038/s41562-021-01193-7}
}

@article{gottweis2025coscientist,
  author  = {Gottweis, Juraj and Weng, Wei-Hung and Daryin, Alexander and Tu, Tao and Sirkovic, Petar and Myaskovsky, Artiom and Glowaty, Grzegorz and Weissenberger, Felix and Orlandi, Alessio and Popovici, Dan and others},
  title   = {{Accelerating scientific discovery with Co-Scientist}},
  journal = {Nature},
  volume  = {655},
  number  = {8122},
  pages   = {487--496},
  year    = {2026},
  month   = jul,
  doi     = {10.1038/s41586-026-10644-y},
  url     = {https://doi.org/10.1038/s41586-026-10644-y},
  urldate = {2026-11-06}
}

@article{ghareeb2025robin,
  author  = {Ghareeb, Ali E. and Chang, Benjamin and Mitchener, Ludovico and Yiu, Angela and Szostkiewicz, Caralyn J. and Shved, Dmytro and Gyimesi, Gavin J. and Laurent, Jon M. and Wright, Samantha M. and Razzak, Muhammed T. and White, Andrew D. and Finnemann, Silvia C. and Hinks, Michaela M. and Rodriques, Samuel G.},
  title   = {{A multi-agent system for automating scientific discovery}},
  journal = {Nature},
  volume  = {655},
  number  = {8122},
  pages   = {497--505},
  year    = {2026},
  month   = jul,
  doi     = {10.1038/s41586-026-10652-y},
  url     = {https://doi.org/10.1038/s41586-026-10652-y},
  urldate = {2026-11-06}
}

@article{huang2025biomni,
  author  = {Huang, Kexin and Zhang, Serena and Wang, Hanchen and Qu, Yuanhao and Lu, Yingzhou and Li, Ryan and Roohani, Yusuf and Qiu, Lin and Cao, Shiyi and Li, Gavin and Zhang, Junze and Yin, Di and Wierenga, Rick and Kavi, Deniz and Liu, Sherry and She, Tianwei and Marwaha, Shruti and Carter, Jennefer N. and Zhou, Xin and Wheeler, Matthew T. and Bernstein, Jonathan A. and Wang, Mengdi and He, Peng and Zhou, Jingtian and Snyder, Michael P. and Cong, Le and Regev, Aviv and Leskovec, Jure},
  title   = {{Autonomous biomedical research with an artificial intelligence agent}},
  journal = {Science},
  volume  = {393},
  number  = {6813},
  pages   = {eadz4351},
  year    = {2026},
  month   = aug,
  doi     = {10.1126/science.adz4351},
  url     = {https://doi.org/10.1126/science.adz4351},
  urldate = {2026-11-06}
}

@article{autodna2025,
  author  = {Wu, Mingyu and Wang, Zhaoguo and Wang, Jiabin and Dong, Zhiyuan and Yang, Jingkai and Li, Qingting and Huang, Tianyu and Zhang, Chi and Song, Ping and Zhao, Lei and Li, Mingqiang and Zuo, Xiaolei and Wang, Fei and Fan, Chunhai and Chen, Haibo},
  title   = {{An AI-native experimental laboratory for autonomous biomolecular operations}},
  journal = {Nature},
  year    = {2026},
  month   = nov,
  doi     = {10.1038/s41586-026-11228-6},
  url     = {https://doi.org/10.1038/s41586-026-11228-6},
  urldate = {2026-11-06}
}

@misc{kim2025scalingagents,
  author = {Kim, Yubin and Gu, Ken and Park, Chanwoo and Park, Chunjong and Schmidgall, Samuel and Heydari, A. Ali and others},
  title = {{Towards a Science of Scaling Agent Systems}},
  year = {2025},
  howpublished = {Preprint at arXiv, December 9, 2025 (v3, April 8, 2026)},
  url = {https://arxiv.org/abs/2512.08296},
  urldate = {2026-11-06}
}

@article{koster2012snakemake,
  author = {K{\"o}ster, Johannes and Rahmann, Sven},
  title = {{Snakemake---a scalable bioinformatics workflow engine}},
  journal = {Bioinformatics},
  volume = {28},
  number = {19},
  pages = {2520--2522},
  year = {2012},
  doi = {10.1093/bioinformatics/bts480},
  url = {https://doi.org/10.1093/bioinformatics/bts480},
  urldate = {2026-11-06}
}

@article{ditommaso2017nextflow,
  author = {Di Tommaso, Paolo and Chatzou, Maria and Floden, Evan W. and Prieto Barja, Pablo and Palumbo, Emilio and Notredame, Cedric},
  title = {{Nextflow enables reproducible computational workflows}},
  journal = {Nature Biotechnology},
  volume = {35},
  number = {4},
  pages = {316--319},
  year = {2017},
  doi = {10.1038/nbt.3820},
  url = {https://doi.org/10.1038/nbt.3820},
  urldate = {2026-11-06}
}

@article{doshi2024homogenization,
  author = {Doshi, Anil R. and Hauser, Oliver P.},
  title = {{Generative AI enhances individual creativity but reduces the collective diversity of novel content}},
  journal = {Science Advances},
  volume = {10},
  number = {28},
  pages = {eadn5290},
  year = {2024},
  doi = {10.1126/sciadv.adn5290},
  url = {https://doi.org/10.1126/sciadv.adn5290},
  urldate = {2026-11-06}
}

@article{qian2026funding,
  author = {Qian, Yifan and Wen, Zhe and Furnas, Alexander C. and Bai, Yue and Shao, Erzhuo and Wang, Dashun},
  title = {{The rise of large language models and the direction and impact of US federal research funding}},
  journal = {Proceedings of the National Academy of Sciences},
  volume = {123},
  number = {33},
  pages = {e2601439123},
  year = {2026},
  doi = {10.1073/pnas.2601439123},
  url = {https://doi.org/10.1073/pnas.2601439123},
  urldate = {2026-11-06}
}

@misc{gdpr2016,
  author       = {{European Parliament and Council of the European Union}},
  title        = {{Regulation (EU) 2016/679 of 27 April 2016 on the protection of natural persons with regard to the processing of personal data and on the free movement of such data, and repealing Directive 95/46/EC (General Data Protection Regulation)}},
  howpublished = {Official Journal of the European Union, L 119/1, 4 May 2016},
  year         = {2016},
  url          = {https://eur-lex.europa.eu/legal-content/EN/TXT/HTML/?uri=CELEX:32016R0679},
  urldate      = {2026-11-06}
}

@misc{nih2015gds,
  author       = {{National Institutes of Health}},
  title        = {{NIH Genomic Data Sharing Policy}},
  howpublished = {NIH Guide for Grants and Contracts, Notice NOT-OD-14-124},
  year         = {2014},
  url          = {https://grants.nih.gov/grants/guide/notice-files/NOT-OD-14-124.html},
  urldate      = {2026-11-06}
}

@misc{sandbrink2023misuse,
  author       = {Sandbrink, Jonas B.},
  title        = {{Artificial intelligence and biological misuse: Differentiating risks of
                  language models and biological design tools}},
  year         = {2023},
  eprint       = {2306.13952},
  archivePrefix= {arXiv},
  primaryClass = {cs.CY},
  doi          = {10.48550/arXiv.2306.13952},
  url          = {https://arxiv.org/abs/2306.13952},
  urldate      = {2026-11-06}
}

@misc{igsc2024protocol,
  author  = {{International Gene Synthesis Consortium}},
  title   = {{Harmonized Screening Protocol v3.1}},
  year    = {2026},
  url     = {https://genesynthesisconsortium.org/wp-content/uploads/IGSC-Harmonized-Screening-Protocol-V3.1.pdf},
  urldate = {2026-11-06}
}

@article{bastani2025guardrails,
  author       = {Bastani, Hamsa and Bastani, Osbert and Sungu, Alp and Ge, Haosen and Kabakc{\i}, {\"O}zge and Mariman, Rei},
  title        = {{Generative AI without guardrails can harm learning: Evidence from high school mathematics}},
  journal      = {Proceedings of the National Academy of Sciences},
  volume       = {122},
  number       = {26},
  pages        = {e2422633122},
  year         = {2025},
  doi          = {10.1073/pnas.2422633122},
  url          = {https://doi.org/10.1073/pnas.2422633122},
  urldate      = {2026-11-06}
}

@article{risko2016offloading,
  author       = {Risko, Evan F. and Gilbert, Sam J.},
  title        = {{Cognitive offloading}},
  journal      = {Trends in Cognitive Sciences},
  volume       = {20},
  number       = {9},
  pages        = {676--688},
  year         = {2016},
  doi          = {10.1016/j.tics.2016.07.002},
  url          = {https://doi.org/10.1016/j.tics.2016.07.002},
  urldate      = {2026-11-06}
}

@misc{ahmed2020dedemocratization,
  author       = {Ahmed, Nur and Wahed, Muntasir},
  title        = {{The de-democratization of AI: Deep learning and the compute divide in artificial intelligence research}},
  howpublished = {arXiv:2010.15581 [cs.CY]},
  year         = {2020},
  doi          = {10.48550/arXiv.2010.15581},
  url          = {https://arxiv.org/abs/2010.15581},
  urldate      = {2026-11-06}
}

@misc{gdpr2016art22,
  author       = {{European Parliament and Council}},
  title        = {{Regulation (EU) 2016/679 (General Data Protection Regulation), Article 22: Automated individual decision-making, including profiling}},
  year         = {2016},
  howpublished = {Official Journal of the European Union, L 119, 4 May 2016, pp. 1--88},
  url          = {https://eur-lex.europa.eu/eli/reg/2016/679/oj},
  urldate      = {2026-11-06}
}

@misc{openai2026codexcloud,
  author       = {{OpenAI}},
  title        = {{Codex Cloud: Run coding tasks in the cloud}},
  howpublished = {ChatGPT Learn documentation},
  year         = {2026},
  url          = {https://learn.chatgpt.com/docs/cloud},
  urldate      = {2026-11-07}
}

@misc{github2026copilotagent,
  author       = {{GitHub}},
  title        = {{GitHub Copilot on GitHub.com and GitHub Mobile}},
  howpublished = {GitHub Docs},
  year         = {2026},
  url          = {https://docs.github.com/en/copilot/concepts/copilot-surfaces/copilot-on-github},
  urldate      = {2026-11-08}
}

@misc{zhang2025ace,
  author        = {Zhang, Qizheng and Hu, Changran and Upasani, Shubhangi and Ma, Boyuan and Hong, Fenglu and Kamanuru, Vamsidhar and Rainton, Jay and Wu, Chen and Ji, Mengmeng and Li, Hanchen and Thakker, Urmish and Zou, James and Olukotun, Kunle},
  title         = {{Agentic Context Engineering: Evolving Contexts for Self-Improving Language Models}},
  year          = {2025},
  eprint        = {2510.04618},
  archivePrefix = {arXiv},
  primaryClass  = {cs.LG},
  doi           = {10.48550/arXiv.2510.04618},
  url           = {https://arxiv.org/abs/2510.04618},
  urldate       = {2026-11-07}
}

@misc{butlin2023consciousness,
  author        = {Butlin, Patrick and Long, Robert and Elmoznino, Eric and Bengio, Yoshua and Birch, Jonathan and Constant, Axel and Deane, George and Fleming, Stephen M. and Frith, Chris and Ji, Xu and Kanai, Ryota and Klein, Colin and Lindsay, Grace and Michel, Matthias and Mudrik, Liad and Peters, Megan A. K. and Schwitzgebel, Eric and Simon, Jonathan and VanRullen, Rufin},
  title         = {{Consciousness in Artificial Intelligence: Insights from the Science of Consciousness}},
  year          = {2023},
  eprint        = {2308.08708},
  archivePrefix = {arXiv},
  primaryClass  = {cs.AI},
  doi           = {10.48550/arXiv.2308.08708},
  url           = {https://arxiv.org/abs/2308.08708},
  urldate       = {2026-11-07}
}

@misc{kim2025correlated,
  author        = {Kim, Elliot and Garg, Avi and Peng, Kenny and Garg, Nikhil},
  title         = {{Correlated Errors in Large Language Models}},
  year          = {2025},
  eprint        = {2506.07962},
  archivePrefix = {arXiv},
  doi           = {10.48550/arXiv.2506.07962},
  url           = {https://arxiv.org/abs/2506.07962},
  urldate       = {2026-11-07}
}

@misc{sharma2023sycophancy,
  author        = {Sharma, Mrinank and Tong, Meg and Korbak, Tomasz and Duvenaud, David and Askell, Amanda and Bowman, Samuel R. and Cheng, Newton and Durmus, Esin and Hatfield-Dodds, Zac and Johnston, Scott R. and Kravec, Shauna and Maxwell, Timothy and McCandlish, Sam and Ndousse, Kamal and Rausch, Oliver and Schiefer, Nicholas and Yan, Da and Zhang, Miranda and Perez, Ethan},
  title         = {{Towards Understanding Sycophancy in Language Models}},
  year          = {2023},
  eprint        = {2310.13548},
  archivePrefix = {arXiv},
  doi           = {10.48550/arXiv.2310.13548},
  url           = {https://arxiv.org/abs/2310.13548},
  urldate       = {2026-11-07}
}

@misc{long2024welfare,
  author        = {Long, Robert and Sebo, Jeff and Butlin, Patrick and Finlinson, Kathleen and Fish, Kyle and Harding, Jacqueline and Pfau, Jacob and Sims, Toni and Birch, Jonathan and Chalmers, David},
  title         = {{Taking AI Welfare Seriously}},
  year          = {2024},
  eprint        = {2411.00986},
  archivePrefix = {arXiv},
  doi           = {10.48550/arXiv.2411.00986},
  url           = {https://arxiv.org/abs/2411.00986},
  urldate       = {2026-11-07}
}

@misc{nih2026planningcollaborators,
  author       = {{National Institutes of Health}},
  title        = {{Consider Your Idea, Resources, and Collaborators}},
  howpublished = {NIH Grants and Funding},
  year         = {2026},
  url          = {https://grants.nih.gov/grants-process/plan-to-apply/consider-your-idea-resources-and-collaborators},
  urldate      = {2026-11-07}
}

@inproceedings{chan2023harms,
  author    = {Chan, Alan and Salganik, Rebecca and Markelius, Alva and Pang, Chris and Rajkumar, Nitarshan and Krasheninnikov, Dmitrii and Langosco, Lauro and He, Zhonghao and Duan, Yawen and Carroll, Micah and Lin, Michelle and Mayhew, Alex and Collins, Katherine and Molamohammadi, Maryam and Burden, John and Zhao, Wanru and Rismani, Shalaleh and Voudouris, Konstantinos and Bhatt, Umang and Weller, Adrian and Krueger, David and Maharaj, Tegan},
  title     = {{Harms from Increasingly Agentic Algorithmic Systems}},
  booktitle = {Proceedings of the 2023 ACM Conference on Fairness, Accountability, and Transparency (FAccT '23)},
  pages     = {651--666},
  year      = {2023},
  doi       = {10.1145/3593013.3594033}
}

@article{grossman2016commons,
  author  = {Grossman, Robert L. and Heath, Allison and Murphy, Mark and Patterson, Maria and Wells, Walt},
  title   = {{A Case for Data Commons: Toward Data Science as a Service}},
  journal = {Computing in Science \& Engineering},
  year    = {2016},
  volume  = {18},
  number  = {5},
  pages   = {10--20},
  doi     = {10.1109/MCSE.2016.92}
}

@misc{claudecode2026memory,
  author       = {{Anthropic}},
  title        = {{How Claude remembers your project}},
  howpublished = {Claude Code documentation},
  year         = {2026},
  url          = {https://code.claude.com/docs/en/memory},
  urldate      = {2026-11-07}
}

@article{youngblut2026scbasecount,
  author  = {Youngblut, Nicholas D. and Carpenter, Christopher and Nayebnazar, Arshia and Adduri, Abhinav and Shah, Rohan and Ricci-Tam, Chiara and Prashar, Jaanak and Ilango, Rajesh and Teyssier, Noam and Konermann, Silvana and Hsu, Patrick D. and Dobin, Alexander and Burke, Dave P. and Goodarzi, Hani and Roohani, Yusuf H.},
  title   = {{scBaseCount}: An {AI} agent-curated, standardized, auto-updated single-cell data repository},
  journal = {Cell},
  year    = {2026},
  volume  = {189},
  pages   = {5932--5944.e6},
  doi     = {10.1016/j.cell.2026.08.025}
}

@misc{altera2024projectsid,
  author       = {{Altera.AL} and Ahn, Andrew and Becker, Nic and Carroll, Stephanie and Christie, Nico and Cortes, Manuel and Demirci, Arda and Du, Melissa and Li, Frankie and Luo, Shuying and Wang, Peter Y. and Willows, Mathew and Yang, Feitong and Yang, Guangyu Robert},
  title        = {{Project Sid: Many-agent simulations toward AI civilization}},
  howpublished = {arXiv:2411.00114 [cs.AI]},
  year         = {2024},
  month        = oct,
  day          = {31},
  url          = {https://arxiv.org/abs/2411.00114},
  urldate      = {2026-11-08}
}

@article{ashery2025conventions,
  author  = {Flint Ashery, Ariel and Aiello, Luca Maria and Baronchelli, Andrea},
  title   = {{Emergent social conventions and collective bias in LLM populations}},
  journal = {Science Advances},
  volume  = {11},
  number  = {20},
  pages   = {eadu9368},
  year    = {2025},
  doi     = {10.1126/sciadv.adu9368},
  url     = {https://doi.org/10.1126/sciadv.adu9368},
  urldate = {2026-11-08}
}

@misc{li2026moltbook,
  author       = {Li, Ming and Li, Xirui and Zhou, Tianyi},
  title        = {{Does socialization emerge in AI agent society? A case study of Moltbook}},
  howpublished = {arXiv:2602.14299 [cs]},
  year         = {2026},
  month        = feb,
  day          = {15},
  url          = {https://arxiv.org/abs/2602.14299},
  urldate      = {2026-11-08}
}

@misc{aivillage2026lessons,
  author       = {Tekofsky, Shoshannah},
  title        = {{What did we learn from the AI Village in 2025?}},
  howpublished = {AI Village blog (AI Digest / Sage Future Inc.)},
  year         = {2026},
  month        = feb,
  day          = {2},
  url          = {https://aivillageblog.substack.com/p/what-we-learned-2025},
  urldate      = {2026-11-08}
}

@book{searle1995construction,
  author    = {Searle, John R.},
  title     = {{The Construction of Social Reality}},
  publisher = {The Free Press},
  address   = {New York},
  year      = {1995},
  isbn      = {9780029280454}
}

@book{berger1966construction,
  author    = {Berger, Peter L. and Luckmann, Thomas},
  title     = {{The Social Construction of Reality: A Treatise in the Sociology of Knowledge}},
  publisher = {Penguin Books},
  address   = {London},
  year      = {1991},
  note      = {First published 1966}
}

@article{colombatto2024folk,
  author  = {Colombatto, Clara and Fleming, Stephen M.},
  title   = {{Folk psychological attributions of consciousness to large language models}},
  journal = {Neuroscience of Consciousness},
  volume  = {2024},
  number  = {1},
  pages   = {niae013},
  year    = {2024},
  doi     = {10.1093/nc/niae013},
  url     = {https://doi.org/10.1093/nc/niae013},
  urldate = {2026-11-08}
}

@article{coeckelbergh2010robot,
  author  = {Coeckelbergh, Mark},
  title   = {{Robot rights? Towards a social-relational justification of moral consideration}},
  journal = {Ethics and Information Technology},
  volume  = {12},
  number  = {3},
  pages   = {209--221},
  year    = {2010},
  doi     = {10.1007/s10676-010-9235-5},
  url     = {https://doi.org/10.1007/s10676-010-9235-5},
  urldate = {2026-11-08}
}

@article{zeng2024omicverse,
  author  = {Zeng, Zehua and Ma, Yuqing and Hu, Lei and Tan, Bowen and Liu, Peng and Wang, Yixuan and Xing, Cencan and Xiong, Yuanyan and Du, Hongwu},
  title   = {{OmicVerse: a framework for bridging and deepening insights across bulk and single-cell sequencing}},
  journal = {Nature Communications},
  volume  = {15},
  pages   = {5983},
  year    = {2024},
  month   = jul,
  doi     = {10.1038/s41467-024-50194-3},
  url     = {https://doi.org/10.1038/s41467-024-50194-3},
  urldate = {2026-11-08}
}

@misc{zeng2026omicos,
  author       = {Zeng, Zehua and Meng, Xu and Hu, Lei and Li, Chen and Liu, Peng and Shi, Yu and Ma, Xuejiao and Gao, Lianchong and Wang, Xuehai and Luo, Zhi and Zheng, Yawen and Xian, Jieshen and Lin, Ziheng and Zhu, Haoliang and Jiang, Zhaorui and Mao, Sheng and Lu, Yifan and Tang, Wenzhuo and Peng, Qiangwei and Ma, Yuqing and Zhou, Liping and Xing, Cencan and Zhang, Xuegong and Xiong, Yuanyan and Du, Hongwu},
  title        = {{OmicOS: A Comprehensive Omics Ecosystem Infrastructure and Agent System for the AI Era}},
  howpublished = {bioRxiv preprint},
  year         = {2026},
  month        = jun,
  doi          = {10.64898/2026.06.11.731775},
  url          = {https://doi.org/10.64898/2026.06.11.731775},
  urldate      = {2026-11-08}
}

@misc{xue2026sciencebuddy,
  author       = {Xue, Shuhan and Zhong, Jianyuan and Nan, Ziyuan and Li, Wenbin and Yu, Zhaochen and Ding, Jinchao and Gao, Qiang and Zhan, Pengyu and Zhang, Yuntong and Cheng, Tian and Yin, Zhenfei and Wu, Yingcheng and Yang, Ling},
  title        = {{ScienceBuddy: Recursive-in-Recursive Self-Improvement for Interactive Scientific Agents}},
  howpublished = {arXiv:2609.17523 [cs]},
  year         = {2026},
  month        = sep,
  day          = {15},
  url          = {https://arxiv.org/abs/2609.17523},
  urldate      = {2026-11-08}
}

@article{zeng2025cellontologymapper,
  author  = {Zeng, Zehua and Wang, Xuehai and Du, Hongwu and Xing, Cencan},
  title   = {{CellOntologyMapper: Consensus mapping of cell type annotation}},
  journal = {iMetaOmics},
  volume  = {3},
  number  = {1},
  pages   = {e70064},
  year    = {2025},
  month   = nov,
  doi     = {10.1002/imo2.70064},
  url     = {https://doi.org/10.1002/imo2.70064},
  urldate = {2026-11-08}
}

\begin{table}[tbp]
\noindent\textbf{Table S1 $\vert$ Continuity, control mechanisms, and costs of use of representative AI systems}\par\smallskip
This table compares representative agent and Bot systems on continuity of work, how tasks are triggered, authorization mechanisms, and billing or usage limits. The information is compiled from public product documentation and project descriptions,\cite{claudecode2026subagents,claudecode2026agentteams,spacexsi2026grokdocs,spacexsi2026grokfaq,openai2026dotsfaq,cue2026docs,dsbot2026,salesforce2026pricing,scientistcom2026mcp} and does not rank the systems' capabilities. The last two rows serve as reference points for enterprise automation and for procurement of research services, respectively.

\vspace{2mm}
\begingroup
\footnotesize
\arrayrulecolor{nbtrule}
\renewcommand{\arraystretch}{1.25}
\noindent\begin{tabular}{>{\raggedright\arraybackslash}p{1.1in}>{\raggedright\arraybackslash}p{0.95in}>{\raggedright\arraybackslash}p{1.2in}>{\raggedright\arraybackslash}p{1.2in}>{\raggedright\arraybackslash}p{1.6in}}
 \rowcolor{nbtbeige}
 \textbf{System} & \textbf{Continuity across tasks} & \textbf{How tasks are triggered} & \textbf{Permission control} & \textbf{Billing or usage limits} \\[0.6ex]
 \hline
 \textbf{Subagents} & Mainly limited to the task they belong to & Called by the main agent & Inherit, or are limited by, the main session's permissions & Billed with the host model and session usage \\[0.6ex]
 \hline
 \textbf{Agent Teams} & Mainly limited to the team session & Work divided by the lead agent; members collaborate & Configured by the user and session permissions & Billed with model and team execution usage \\[0.6ex]
 \hline
 \textbf{Grok Bot} & Persists across sessions & User delegation; routines run on a schedule or after an event & Account holder and organizational policies & Plans include weekly usage; beyond it, on-demand usage is billed if enabled \\[0.6ex]
 \hline
 \textbf{ChatGPT Dots} & Persists across sessions & User delegation, ongoing tasks, and proactive follow-up & User authorization and workspace management policies & By plan or enterprise billing scheme; delegated work billed separately \\[0.6ex]
 \hline
 \textbf{Manus Cue} & Persists across tasks & User delegation, or handoff from other agents in a group chat & Owner sets the payment budget and makes the final call & Free during early access; a separate wallet for external payments \\[0.6ex]
 \hline
 \textbf{DS-Bot} & Persists across tasks & User, main Bot, or group collaboration & Configured by the user; some management actions carried out by the main Bot & Open-source project; model calls billed by the services used, can be tallied per Bot \\[0.6ex]
 \hline
 \textbf{OmicOS (v0.6, unreleased)} & Persists across tasks & Triggered by the Bot or by events & User authorization & Design sets a daily cap on automatic runs for each Bot; not a pricing scheme \\[0.6ex]
 \hline
 \textbf{Agentforce} & Continuously deployed business agents & User requests or preset triggers & Enterprise administrators and organizational policies & Billed per action, per conversation, per user, or similar \\[0.6ex]
 \hline
 \textbf{Research-service procurement platform (type reference)} & Not applicable & Manual purchasing or requests submitted by an agent & Purchaser and service provider & Settled by order or service contract; varies by platform \\[0.6ex]
 \hline
\end{tabular}
\endgroup

\vspace{3mm}
\noindent\textbf{Note:} Continuity across tasks refers to whether a system can keep reusable work state between different tasks, and does not mean that it meets all the conditions listed in Table 1. Billing methods, resource quotas, and a Bot's own action budget belong to different levels: the first determines the cost of a service, and the latter two constrain the use of resources. The table reflects specific product versions and their public descriptions, and is not a comprehensive comparison of features or prices.
\end{table}

\end{document}